\documentclass[pdflatex,sn-mathphys-num]{sn-jnl}

\usepackage{graphicx}%
\usepackage{multirow}%
\usepackage{amsmath,amssymb,amsfonts}%
\usepackage{amsthm}%
\usepackage{mathrsfs}%
\usepackage{appendix}%
\usepackage{xcolor}%
\usepackage{textcomp}%
\usepackage{manyfoot}%
\usepackage{booktabs}%
\usepackage{algorithm}%
\usepackage{algorithmicx}%
\usepackage{algpseudocode}%
\usepackage{listings}%
\usepackage{graphicx}%
\usepackage{hyperref}%
\usepackage{cleveref}%
\usepackage{chngcntr}%
\usepackage{subcaption}
\usepackage{soul}
\crefname{table}{Table}{Tables}
\Crefname{table}{Table}{Tables}
\crefname{figure}{Figure}{Figures}
\Crefname{figure}{Figure}{Figures}
\crefname{appendix}{Appendix}{Appendices}
\Crefname{appendix}{Appendix}{Appendices}

\theoremstyle{thmstyleone}%
\theoremstyle{thmstyletwo}%

\theoremstyle{thmstylethree}%

\begin{document}

\title[Interactive Symbolic Regression through Offline Reinforcement Learning: A Co-Design Framework]{Interactive Symbolic Regression through Offline Reinforcement Learning: A Co-Design Framework}


\author[1,2]{\fnm{Yuan} \sur{Tian}}\email{tian\_yuan4@ctg.com.cn}

\author[1,3]{\fnm{Wenqi} \sur{Zhou}}\email{wenzhou@student.ethz.ch}

\author[3]{\fnm{Michele} \sur{Viscione}}\email{michele.viscione@epfl.ch}

\author[1,3]{\fnm{Hao} \sur{Dong}}\email{hao.dong@ibk.baug.ethz.ch}

\author[1]{\fnm{David} \sur{S. Kammer}}\email{dkammer@ethz.ch}

\author*[3]{\fnm{Olga} \sur{Fink}}\email{olga.fink@epfl.ch}

\affil[1]{\orgdiv{Institute for Building Materials}, \orgname{ETH Z\"urich}, \orgaddress{\street{Laura-Hezner-Weg 7}, \city{Z\"urich}, \postcode{8093}, \state{Z\"urich}, \country{Switzerland}}}

\affil[2]{\orgname{China Yangtze Power Co.,Ltd}, \orgaddress{\street{No.1 Xiba jianshe Road}, \city{Yichang}, \postcode{443000}, \state{Hubei}, \country{China}}}

\affil[3]{\orgdiv{Laboratory of Intelligent Maintenance and Operations Systems}, \orgname{EPFL}, \orgaddress{\street{Station 18}, \city{Lausanne}, \postcode{1015}, \state{Lausanne}, \country{Switzerland}}}


\abstract{Symbolic Regression (SR) holds great potential for uncovering underlying mathematical and physical relationships from observed data. However, the vast combinatorial space of possible expressions poses significant challenges for both online search methods and pre-trained transformer models. Additionally, current state-of-the-art approaches typically do not consider the integration of domain experts' prior knowledge and do not support iterative interactions with the model during the equation discovery process. To address these challenges, we propose the Symbolic Q-network (Sym-Q), an advanced interactive framework for large-scale symbolic regression. Unlike previous large-scale transformer-based SR approaches, Sym-Q leverages reinforcement learning without relying on a transformer-based decoder. This formulation allows the agent to learn through offline reinforcement learning using any type of tree encoder, enabling more efficient training and inference. Furthermore, we propose a co-design mechanism, where the reinforcement learning-based Sym-Q facilitates effective interaction with domain experts at any stage of the equation discovery process. Users can dynamically modify generated nodes of the expression, collaborating with the agent to tailor the mathematical expression to best fit the problem and align with the assumed physical laws, particularly when there is prior partial knowledge of the expected behavior. Our experiments demonstrate that the pre-trained Sym-Q surpasses existing SR algorithms on the challenging SSDNC benchmark. Moreover, we experimentally show on real-world cases that its performance can be further enhanced by the interactive co-design mechanism, with Sym-Q achieving  greater  performance gains than other state-of-the-art models. Our reproducible code is available at \href{https://github.com/EPFL-IMOS/Sym-Q}{https://github.com/EPFL-IMOS/Sym-Q}.}


\maketitle

\section{Introduction}\label{sec: introduction}
Symbolic regression is a powerful form of regression analysis that searches the space of mathematical expressions to find the expression that best fits an observed dataset. Unlike traditional regression models that fit data to pre-specified equations, symbolic regression can discover the underlying equations or relationships between variables. This capability can lead to a deeper understanding of the inherent structure and dynamics of the underlying processes. It is particularly important in fields where the relationships between variables are complex and not well understood, as it provides a tool to uncover the form of the relationship without prior assumptions. Recently, symbolic regression has been instrumental in uncovering new relationships, such as astrophysical scaling relations~\cite{wadekar2022augmenting} and analytical models of exoplanet transit spectroscopy~\cite{ matchev2022analytical}. However, a significant challenge in symbolic regression is its inherent combinatorial complexity. This complexity grows with the length of the symbolic expressions, making it a computationally demanding NP-hard problem~\cite{li2022transformer,virgolin2022symbolic}.

To address this challenge, researchers have made significant advancements in symbolic regression mainly in two methodological directions: online search techniques and transformer-based models trained on large-scale datasets. Online search methods aim to identify mathematical expressions that best describe the data by efficiently exploring the solution space. One of the most widely used methods in online search for symbolic regression is Genetic Programming (GP)~\cite{forrest1993genetic, schmidt2009distilling, blkadek2019solving}. GP iteratively evolves successive generations of mathematical expressions to approximate observed data by applying operations such as selection, crossover, and mutation. Another effective method in online search is reinforcement learning (RL), which takes a different approach to optimizing expressions by learning policies that guide the search process. For example, \citet{petersen2019deep} introduced Deep Symbolic Regression (DSR), which utilizes a risk-seeking policy gradient to optimize expressions using the normalized root-mean-square error as a reward signal. While promising, this approach has a significant drawback: it requires searching for every expression from scratch, making it computationally expensive and time-consuming. Moreover, combining GP and RL can leverage the strengths of both strategies. \citet{mundhenk2021symbolic} introduced a neural-guided component to initialize the starting population for a random-restart GP process. This hybrid approach enables the model to progressively learn and refine better starting populations,  resulting in significant performance improvements over traditional methods.

However, these online search methods often require training a new model for each specific expression, resulting in limited generalization and substantial computational demands. To address these limitations, recent research has shifted focus toward transformer decoder-based models~\cite{li2022transformer, vastl2022symformer, kamienny2022end, meidani2023snip, li2025mmsr} to construct the output expression tree. Recent studies investigated different points encoder~\cite{li2022transformer}, contrastive loss~\cite{li2022transformer} and fusion strategies~\cite{meidani2023snip, li2025mmsr}. These models leverage powerful attention mechanisms and are trained on large-scale datasets, enabling them to effectively learn and understand different patterns. This allows them to autonomously generate plausible mathematical expressions in an auto-regressive manner, dynamically adapting to the input data. Transformer-based approaches also excel at capturing  long-range dependencies within data, which is crucial for tackling complex symbolic regression tasks. Recently, pre-trained transformer models have been utilized to enhance the online search process by integrating advanced techniques like Monte Carlo Tree Search~\cite{li2024discovering} and Monte Carlo Sampling~\cite{li2024generative} into the pipeline, further improving efficiency and performance.

While these transformer decoder-based methods~\cite{li2022transformer, vastl2022symformer, kamienny2022end} demonstrate strong potential in advancing symbolic regression—particularly in terms of scalability and generalizability—they also rely on a traditional teacher-forcing training paradigm. In this setup, the model’s next-token predictions are directly conditioned on the ground-truth tokens, creating an exposure bias between training and inference. This not only introduces a gap that can lead to aggregated error during inference, but also allows gradients to propagate across multiple tokens, resulting in overly optimistic updates that may not reflect real inference dynamics~\cite{bahdanau2016actor,xu2019rethinking,tsai2020order}.

Building on this, the recently proposed Neural Symbolic Regression with Hypothesis (NSRwH)~\cite{bendinelli2023controllable} introduced an innovative framework that incorporates explicit assumptions about the underlying structure of the target expression to predict equations more effectively. This approach employs an additional transformer to process user prompts. However, despite these advancements, the model may not always reliably meet the specific requirements outlined in the input prompts, and its effectiveness remains limited without extensive prior knowledge to guide its predictions.

Inspired by recent efforts to integrate contextual knowledge and insights from human experts, and to address existing limitations, we propose the co-design mechanism in symbolic regression. This approach involves human experts sequentially modifying or adapting expressions based on discrepancies between observed data points and the predicted curve of the proposed expression. Co-design in symbolic regression involves an iterative interaction between a reinforcement learning agent and domain experts. This method has the potential to seamlessly integrate domain expertise with the advanced learning capabilities of Deep Reinforcement Learning (DRL), ensuring both flexibility and intuitiveness.

\begin{figure}[ht!]
    \centering
    \includegraphics[width=0.87\linewidth]{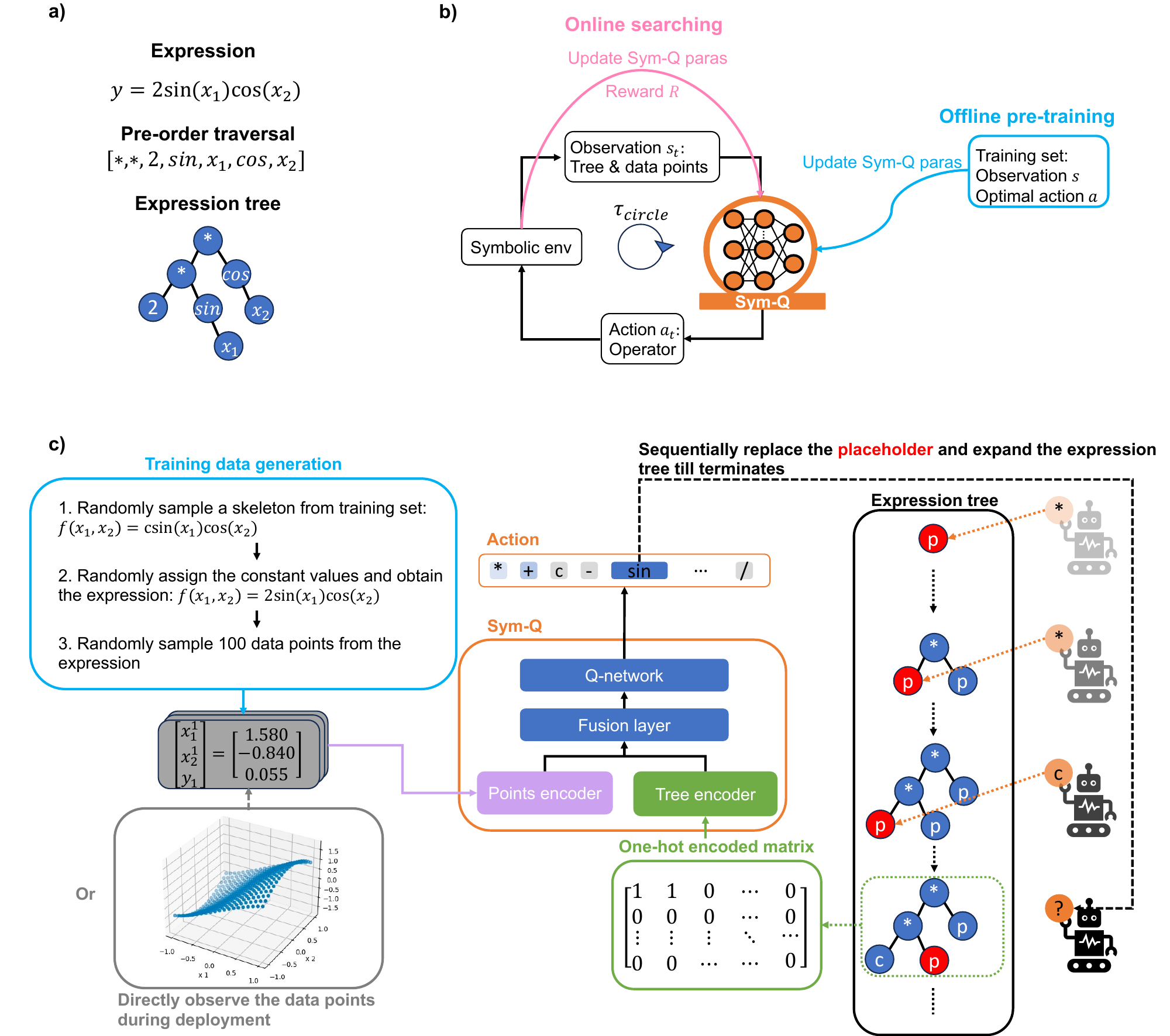}
    \caption{Overview of the proposed framework. \textbf{a).} The expression and its corresponding expression tree. \textbf{b).} The proposed Sym-Q agent supports both offline training with ground truth human knowledge and potentially online searching with reward signals. $\tau_{circle}$ represents the agent trajectories within the symbolic environment. \textbf{c).} The Sym-Q architecture and step-wise decision-making mechanism.
    }
    \label{fig: framework}
\end{figure}

In this study, we propose the Symbolic Q-network (Sym-Q), a novel interactive RL-based method for symbolic regression, as illustrated in \cref{fig: framework}. Sym-Q innovatively incorporates the existing expression tree as part of its observation (\cref{fig: framework}.\textbf{a}). An RL agent then determines the optimal operations to expand and refine this tree until a satisfactory expression is formed (\cref{fig: framework}.\textbf{c}). Unlike existing approaches, Sym-Q can handle the tree representation using various types of encoders, such as transformers or simple recurrent neural networks (RNNs), making it highly flexible and computationally efficient. This flexibility sets Sym-Q apart from other methods, as it does not rely on a transformer-decoder structure to complete expressions. Instead, it leverages the strength of reinforcement learning to guide the agent step-by-step through the process of building an equation, introducing a new paradigm in symbolic regression.

Importantly, Sym-Q supports prompt inputs like predefined expression trees and facilitates interactive design without the need for additional modules, such as extra prompt encoders~\cite{bendinelli2023controllable}. Since the agent makes decisions at the level of individual operations rather than entire sequences, directly modifying the initial expression tree, the integration of prior knowledge is seamlessly preserved throughout the process. The proposed co-design mechanism optimally leverages the capabilities of reinforcement learning and the contextual insights of domain experts. This collaboration enhances the accuracy, relevance, robustness, and meaningfulness of the models, ensuring that the agents align more closely with the true underlying physical processes and dynamics, resulting in more effective and impactful solutions. We demonstrate Sym-Q's superior performance on the challenging SSDNC dataset compared to strong transformer-based baselines. Additionally, we conduct a detailed analysis of fault patterns, examining error distributions and the specific stages at which errors appear during the expression generation process. Moreover, we conduct three experiments to evaluate the effectiveness of the proposed co-design mechanism. The first experiment tests its performance on the SSDNC benchmark. The second experiment involves recovering unseen drift terms in the Feynman dataset by utilizing the original equation as prior knowledge. The third experiment derives an analytic expression for a synthetic transit spectra dataset through an iterative co-design process guided by domain expert insights. The results demonstrate that incorporating prior knowledge significantly enhances expression recovery performance.

\section{Results}\label{sec: results}

\subsection{Training datasets}\label{subsec: training datasets}
For our training dataset, we generated five million mathematical expressions based on $100,000$ pre-defined expression skeletons with up to two independent variables, following the same setup as used in T-JSL~\cite{li2022transformer}. These skeletons were created using the method described by Deep Learning for Symbolic Mathematic~\cite{lample2019deep}, which involves constructing a random unary-binary tree, populating its nodes with operators, and filling its leaves with independent variables or constants. We varied the constants in each expression $50$ times and sampled $100$ random data points for each variation. Further details about the dataset and the corresponding action space are available in \cref{appendix: details of neural networks and hyperparameters}.


\subsection{Evaluation on benchmark datasets}\label{subsec: evaluation on benchmark datasets}
In this study, our primary focus is on generating accurate expression skeletons, which are essential for uncovering the underlying physical relationships. It is important to recognize that while various combinations of operator terms can achieve  a high coefficient of determination $(R^2)$ ~\cite{glantz2001primer}, a widely used metric in the field of symbolic regression, this does not inherently ensure the precision of final predictions or the correctness of the derived  expressions~\cite{li2022transformer}. $R^2$ is defined as:

\begin{equation}
    R^2 = 1 - \frac{\sum_{i=1}^{n}(y_i - \hat{y}_i)^2}{\sum_{i=1}^{n}(y_i - \bar{y})^2} ~,
\end{equation}

where $y_i$ and $\hat{y}_i$ represent the ground-truth and predicted values for point $i$, respectively. $\bar{y}$ is the average of the $y_i$ values across all data points, and $n$ denotes the number of observed data points. Since our focus is on learning the correct expressions, we emphasize evaluating the skeleton recovery rate. The skeleton recovery rate is defined as the percentage of cases where the model fully recovers the correct expression structure, relative to the total number of expressions in the dataset. For the evaluation, we use the challenging SSDNC dataset proposed by \citet{li2022transformer}. This test dataset includes the same skeletons as those generated in the training set but features different numerical coefficients. The SSDNC dataset, currently the most comprehensive of its kind, comprises 963 unseen equations. Further details about the SSDNC dataset are available in \cref{appendix: details of dataset generation and action space}. Since Sym-Q is a large-scale symbolic regression model, we compare its performance with three state-of-the-art (SOTA) transformer-based supervised learning methods, SymbolicGPT~\cite{valipour2021symbolicgpt}, NeSymReS~\cite{biggio2021neural} and T-JSL~\cite{li2022transformer} rather than with other online search-based methods. Sym-Q achieves a recovery rate of \textbf{42.7$\%$} without beam search and improves significantly to \textbf{82.3$\%$} with the implementation of beam search. All comparative methods, including those utilizing beam search, were evaluated under the same configuration with a beam size of 128. Under this setup, our model's inference time is approximately 1 minute per expression with 100 data points, while NeSymReS~\cite{biggio2021neural} needs around 6 minutes. As detailed in \cref{tab: main_result}, our proposed method outperforms state-of-the-art (SOTA) methods by up to $32.0\%$ and achieves a superior average $R^2$ score of \textbf{0.95135}.

Sym-Q achieves superior weighted average $R^2$ fitting accuracy across five well-recognized benchmarks, including Nguyen~\cite{uy2011semantically}, Constant, Keijzer~\cite{keijzer2003improving}, R rationals~\cite{krawiec2013approximating}, and AI-Feynman~\cite{udrescu2020ai}. In addition to the three transformer-based methods previously outlined, we compared our approach with two online search approaches: Deep Symbolic Regression (DSR)~\cite{petersen2019deep} and standard genetic programming-based symbolic regression~\cite{koza1994genetic}. As detailed in \cref{tab:benchmarks}, Sym-Q outperforms all these methods, achieving a weighted average $R^2$ of \textbf{0.95044} across these six benchmarks. Since skeleton recovery rates have not been previously reported for these benchmarks, our analysis focuses solely on the weighted average $R^2$. Further details are provided in \cref{appendix: fitting accuracy on additional benchmarks}. All evaluations and baseline implementations were conducted using the setup described by \citet{li2022transformer}.

\begin{table}[h]
    \caption{Recovery rate of expression skeletons and $R^2$ values on the SSDNC benchmark dataset. All models implemented the beam search strategy with a beam size of 128.}
    \label{tab: main_result}
    \begin{tabular*}{\textwidth}{@{\extracolsep\fill}lcc}
    \toprule
    \textbf{Methods} & \textbf{Recovery rate} & \textbf{$R^2$} \\
    \midrule
    SymbolicGPT & $50.3\%$ & 0.74087 \\
    NeSymReS & $63.4\%$ & 0.85792 \\
    T-JSL & $75.2\%$ & 0.94782 \\
    \hline
    Ours & \textbf{82.3$\%$} & \textbf{0.95135} \\
    \botrule
    \end{tabular*}
\footnotetext{Note: The recovery rate represents the percentage of correctly recovered expression skeletons. $R^2$ is the coefficient of determination for the predicted results.}
\end{table}

Furthermore, to demonstrate scalability, we trained an additional model with three independent variables and evaluated its performance on SRBench against the pretrained model NeSymReS, which also utilizes three input variables. Sym-Q achieved superior equation recovery and higher $R^2$ , as detailed in \cref{appendix: evaluation on srbench}.

\subsection{Evaluation of the Co-design mechanism}\label{Sec:Evaluation of the Co-design mechanism}

To validate the effectiveness of the proposed co-design mechanism in enhancing model performance, we conducted a series of targeted experiments.
\subsubsection*{Co-design's effectiveness}

\begin{figure*}[h!]
    \centering
    \includegraphics[width=0.95\linewidth]{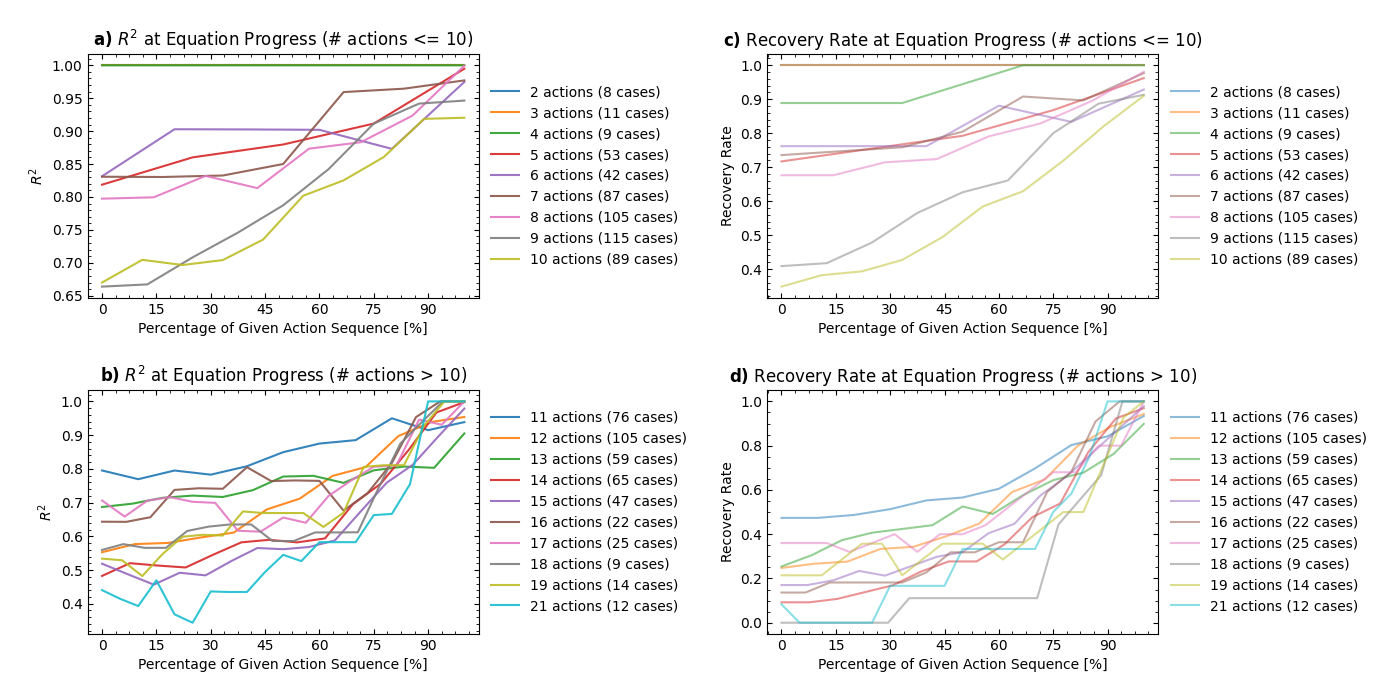}
    \caption{The figures illustrate enhanced  model performance on the SSDNC dataset when partial ground truth of the equation is provided by domain experts ($R^2$ (a, b) and recovery rate (c, d) ). The x-axis represents  the percentage of given action sequences relative to the total. We analyzed equations of varying lengths, excluding categories with fewer than five samples to ensure  statistical validity. Notably, both average $R^2$ and recovery rates increase  as more ground truth sequence steps are incorporated, demonstrating  that domain expert guidance  significantly improves model performance.}
    \label{fig: performance improvement via co-design}
\end{figure*}

In the first experiment, we  evaluated  Sym-Q's co-design mechanism by testing its ability to improve expression recovery rates for unseen mathematical equations using the SSDNC dataset~\cite{li2022transformer}. We  compared its performance  against NeSymReS~\cite{biggio2021neural},  not initially designed  for co-design purposes.  To adapt NeSymReS for the co-design process, known tokens were provided to the decoder, which then predicted  subsequent  tokens auto-regressively until reaching the "end" token. To ensure fairness, expressions were grouped based on the length of their ground truth decision sequences, performance metrics were averaged within these groups. Groups with fewer than five cases were excluded to ensure statistical validity. For each equation with a decision sequence of $n$ actions, we performed a series of $n$ experiments.

\begin{figure}[h]
\centering
\includegraphics[width=\columnwidth]{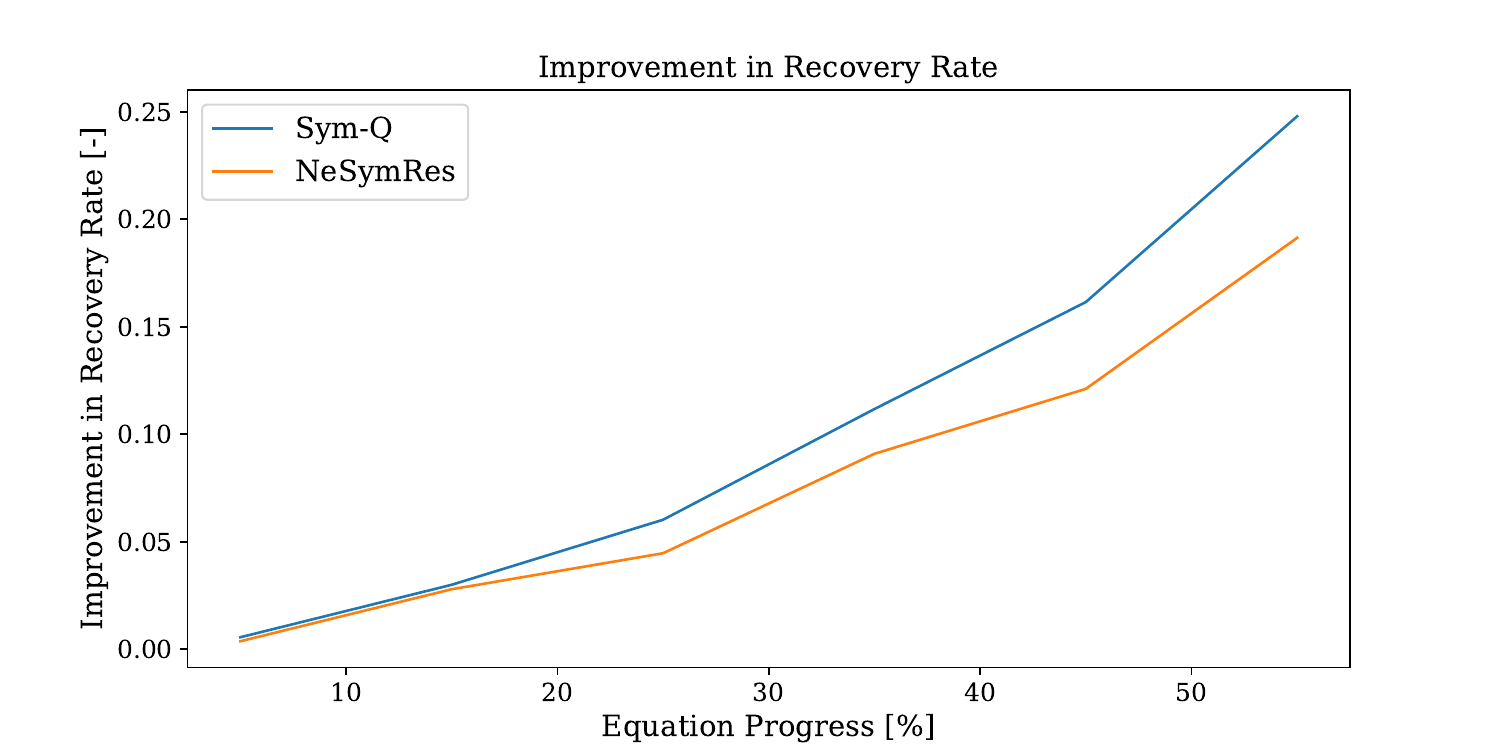}
\caption{Overall performance improvement with respect to the ratio of ground truth information provided for Sym-Q and NeSymReS, showing the co-design impact with up to a maximum of 60\% additional information provided.}
\label{fig:co-design comparison 2}
\end{figure}

Initially, the model was presented with an empty expression tree, corresponding to zero actions taken from the ground truth decision sequence, representing a scenario with no prior expert knowledge. Subsequently, we incrementally incorporated domain knowledge, simulating human interaction or co-design at each step. 
Specifically, in the $i$th experiment, the expression tree included the first $i-1$ actions provided from human expert's ground truth sequence. This setup  allowed us to evaluate the model's performance improvement as more human interaction was integrated. We recorded  performance at each increment and calculated the average  for the respective groups. Trends observed in \cref{fig: performance improvement via co-design} show  a clear enhancement in both $R^2$ and skeleton recovery rate metrics as additional ground truth steps are provided, demonstrating Sym-Q's potential to improve performance through strategic  co-design integration. Moreover, Sym-Q's co-design mechanism outperformed  NeSymReS~\cite{biggio2021neural}. \cref{fig:co-design comparison 2} illustrates  the improvement in recovery rates across different percentages of provided ground truth information, specifically focusing on cases where the initial recovery rate is below 75\%. The results show that while both Sym-Q and NeSymReS improve performance as  more information becomes available, Sym-Q exhibits a steeper and more consistent improvement curve. This indicates  that Sym-Q's design is better suited  to effectively utilize partial information, achieving higher recovery rates  under the co-design mechanism.  For more detailed analysis, please refer to Appendix \ref{appendix: co-design comparison}.

\subsubsection*{Co-design on the Feynman benchmark}
\begin{figure}[h!]
    \centering
    \includegraphics[width=0.95\linewidth]{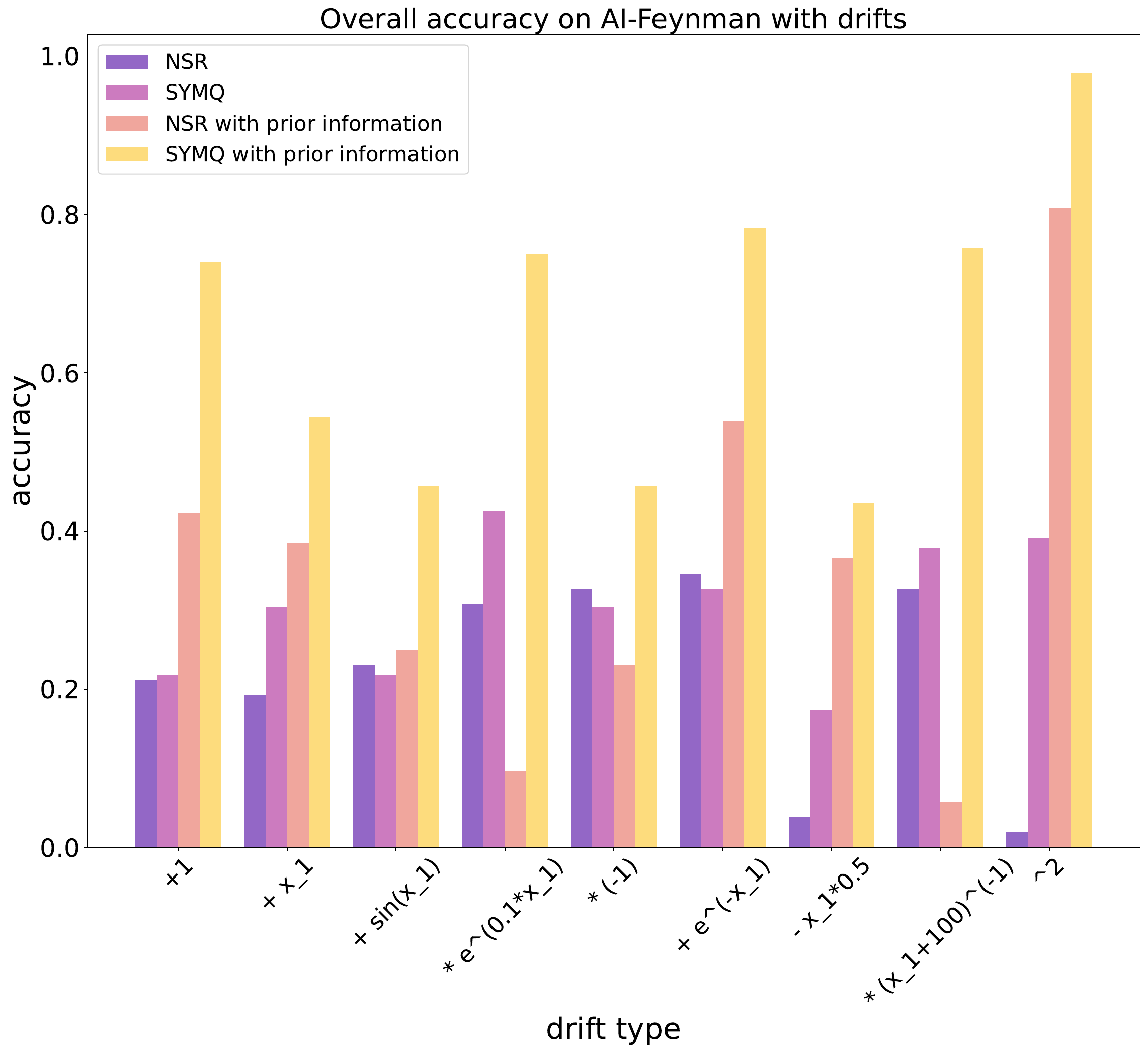}
    \caption{Model accuracy in retrieving different additive and multiplicative drifts over the AI Feynman benchmark. Violet (NeSymRes) and magenta (Sym-Q) bars depict the recovery rate of equations with additional drifts, without giving any prior knowledge. Salmon (NeSymRes) and yellow (Sym-Q) bars refer to the case where the textbook equation is given as a prior and, therefore, only the additional components need to be recovered.}
    \label{fig:drifts_accuracy}
\end{figure}

In the second experiment, we evaluated the model's ability to extend known equations to account for unmodeled dynamics using the Feynman dataset. We selected all equations with three or fewer variables, excluding those exceeding the model's size limits. The experimental setting addresses unmodeled dynamics by aiming to extend "textbook" equations to account for deviations in observations. Specifically, we aim to identify potential "drifts", which are missing multiplicative or additive terms in  known equations that describe the general behavior of observations or underlying laws. For example, additional factors such as air friction may influence a free fall experiment. Using the AI-Feynman dataset, which contains physics-based equations, we evaluated the model's accuracy in retrieving different types of drift, including  multiplicative and additive terms, exponential decay, periodic terms and squared terms of the original equations. The results highlight that the co-design mechanism significantly improves the model's performance in addressing such tasks.
Figure \ref{fig:drifts_accuracy}  demonstrates the effectiveness of Sym-Q in recovering different drift scenarios, where human experts provide the core components of textbook formulas. Both models were evaluated without beam search as it is challenging to decouple model performance from the online search enhancements in this setting. For consistency, we employed the same BFGS setup for both models to determine  the constants of output skeleton.


The performance of Sym-Q and NeSymRes is comparable when prior information is not provided. However, a notable distinction arises when the models is provided with "textbook" equations. In such cases, Sym-Q demonstrates superior capabilities in  co-design contexts leveraging  prior information. Interestingly, while  prior knowledge can sometimes limit  a model’s expressivity and degrade  its performance, as observed with NeSymRes in Figure \ref{fig:drifts_accuracy}, Sym-Q remains unaffected by this limitation. This challenge often arises when out-of-distribution (OOD) priors are provided, where equation skeletons differ structurally from those the model was trained on, as the same equation can often be expressed in multiple forms. Sym-Q’s ability to overcome these constraints underscores its  exceptional  generalization capabilities.

A key  distinction that can partly explain  these differences is NeSymRes's ability to generate  an \textit{end-of-equation} token. While this token  allows the model to decide when an equation is "good enough",  it can also hinder performance by prematurely halting the generation process. In contrast, Sym-Q does not rely on an explicit \textit{end-of-equation} token; instead it uses consistency checks to ensure  that the generated symbolic expressions are complete and contain no unfinished  terms. To prevent NeSymRes from stopping prematurely when provided  with a context equation, we restricted  the model from outputting the \textit{end-of-equation} token as an initial output. Additionally, we implemented the same consistency checks used by Sym-Q as an extra stopping policy, ensuring a fair comparison between the two approaches.

Regarding the drifts, Sym-Q successfully recovered additive and multiplicative exponential factors over $75\% $of the equations, demonstrating its ability to identify unmodeled decaying factors (e.g., resistance or friction) as well as exponentially increasing drifts (e.g., unstable modes in the underneath dynamics). Furthermore, the model effectively recovered additive constant and linear terms, which are often  associated with biased or miscalibrated sensors. Recovery rates ranged from $40\%-80\%$, depending on the relative magnitude of the bias within the data.

\subsubsection*{Co-design on a real use case}  
To further evaluate the versatility of Sym-Q with co-design, we applied  it to derive an analytical expression for the transit radius factor $f$ of a hot-Jupiter exoplanet using atmospheric parameters ($\pi_i$). Building on  the work from \citet{matchev2022analytical}, we tested Sym-Q using  a synthetic transit spectra dataset~\cite{marquez2018supervised}. Applying  the model without co-design, and running it auto-regressively on the unprocessed data, we obtained a solution with a low MSE of $7.6 \times 10^{-4}$; however, the resulting  equation had an $R^2$ value below $0.95$, indicating a significant deviation from the ground truth. This discrepancy is primarily due to the differing variable ranges in the dataset $[10^{-2}, 10^3]$ compared to the training distribution of Sym-Q $[10^{-1}, 10^1]$. While rescaling could mitigate this issue, it would also broaden the range of constant values, complicating the online constants search.

\begin{figure}[th!]
    \centering
    \centering
    \begin{subfigure}[b]{0.45\linewidth}
        \includegraphics[width=\linewidth]{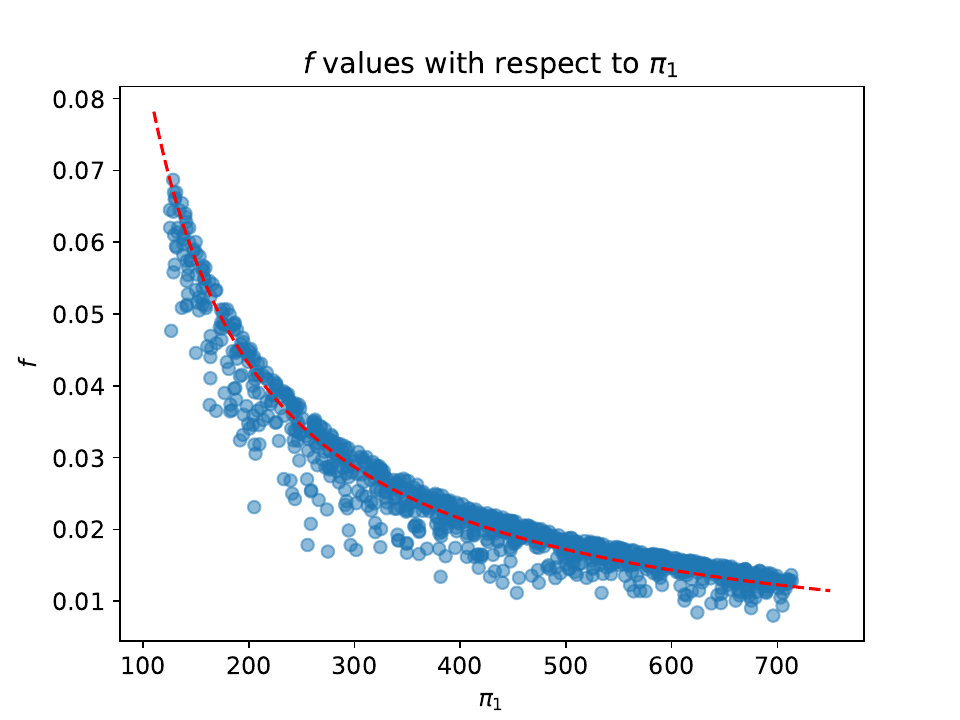}  
        \caption{}
    \end{subfigure} \quad
    \begin{subfigure}[b]{0.45\linewidth}
        \includegraphics[width=\linewidth]{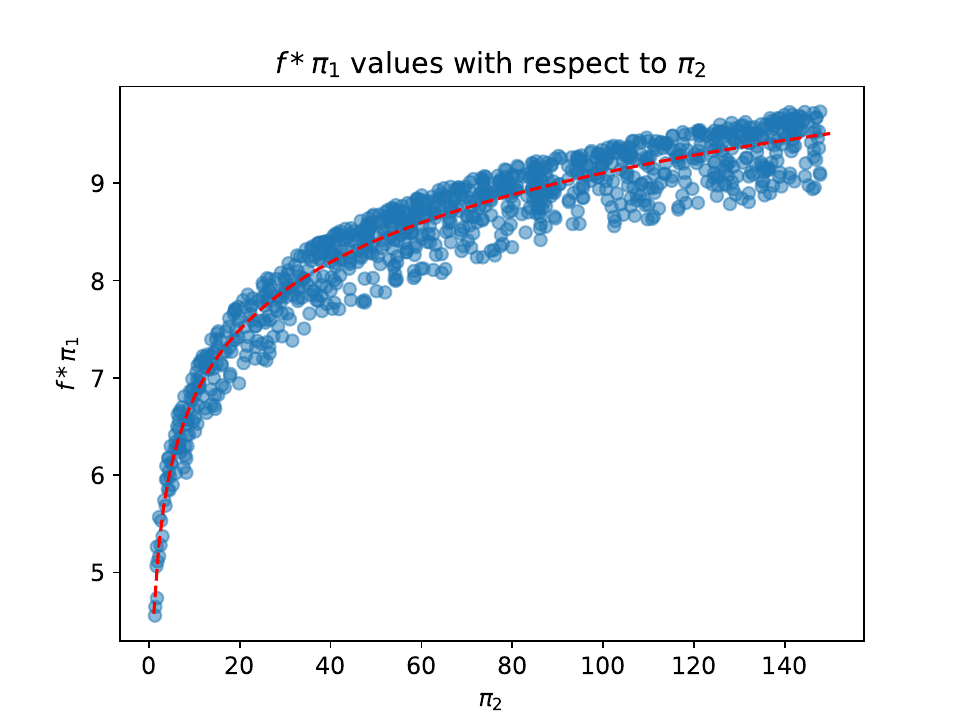}     
        \caption{}
    \end{subfigure} \\
    
    \caption{Correlation between the transit spectra dataset~\cite{marquez2018supervised} entries $\pi_1$, $\pi_2$ and the observation values $f$ and $f\pi_1$. These were used to set up the two priors we used to co-design the analytical expression of $f$.}
    \label{fig:expert_priors}
\end{figure}

Upon further examination of the data, we make the following observations. Firstly we observed that:
\newline

\begin{center}
    (\textbf{I}) $f$ is inversely proportional to $\pi_1$ (\cref{fig:expert_priors}a).
    \newline
\end{center}

Based on this insight, we provide the prior $f = \frac{1}{\pi_1 + c}*(\cdot)$ to the model. This significantly improves the performance, resulting in an MSE of $4 \times 10^{-6}$ and an $R^2$ of $0.974$. We infer that the analytical expression of $f$ takes the form: 

$$f = \frac{1}{\pi_1 + c}*g(\pi_1, \pi_2)$$
  
where $\{\pi_i\}_{i>2}$ are superfluous. Secondly we observe that:
\newline

\begin{center}
    (\textbf{II}) $g \approx f \cdot \pi_1$ scales logarithmically with respect to $\pi_2$ (\cref{fig:expert_priors}b).
    \newline
\end{center}

With this in mind, we propose the prior $f = \frac{1}{\pi_1 + c} * \log(\pi_2 *(\cdot))$ and successfully retrieve the ground truth analytical expression:

$$f(\pi_1, \pi_2, \pi_\text{extra}) = f(\pi_1, \pi_2) = \frac{(c + log(c*sqrt(\pi_1)*\pi_2))}{(c + \pi_1)}$$  

By incorporating astrophysical constants as additional task-specific priors into the BFGS optimization, the online search ultimately converges to the correct closed-form solution:
  
$$f(\pi_1, \pi_2, \pi_\text{extra}) = f(\pi_1, \pi_2) = \frac{\gamma_E + log(2\pi_2\sqrt{\pi\pi_1})}{\pi_1}$$  

This iterative co-design process, supported by domain expert insights, allows Sym-Q to accurately recover the underlying physical equation by strategically guiding the model through informed priors and targeted refinements.

\subsection{Detailed error analysis of agent's decision-making}
\label{subsec: detailed error analysis of agent's decision-making}
To thoroughly evaluate the agent's performance, we conducted a step-wise analysis of its decision-making process using the SSDNC dataset. This section presents key findings, including the error rate trend throughout the expression generation process, the distribution of incorrect selections, and the specific types of errors encountered. Step-wise accuracy was calculated by assessing whether the model's output decision sequence aligned with the expected decision sequence from the SSDNC dataset, based on the correct preceding expression tree encoding and the observed data point.

\begin{figure}[h!]
\centering
\includegraphics[width=0.85\columnwidth]{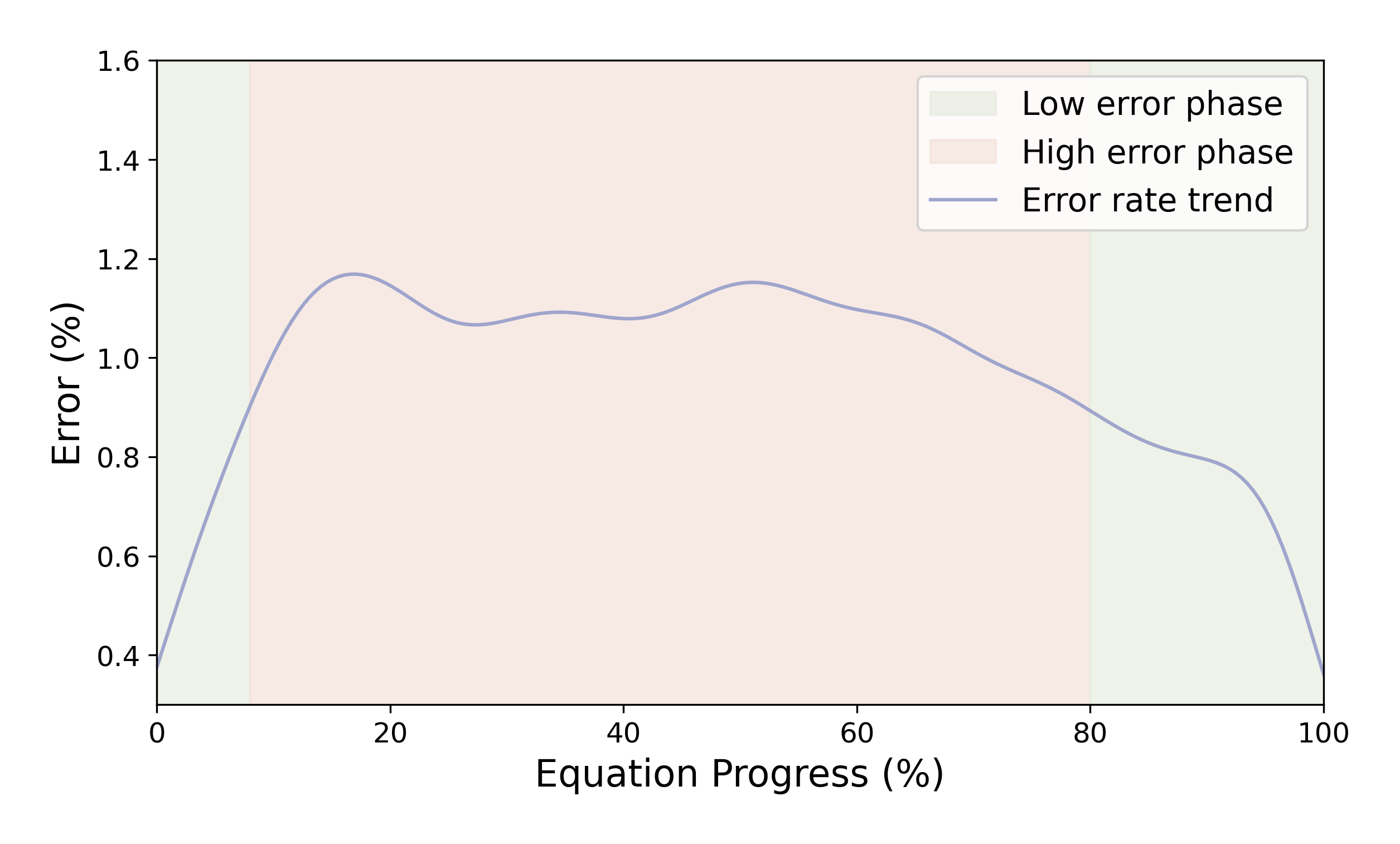}
\caption{Error rate trend throughout the expression generation process. The y-axis displays error rates, and the x-axis represents the percentage of completion of the expression generation process, with $100\%$ indicating a fully formed equation. A distinct pattern is evident: higher error rates occur in the middle stages of the decision-making process, while the initial and final phases exhibit lower error rates.}
\label{depth}
\end{figure}

\begin{figure}[h!]
\centering
\includegraphics[width=1\columnwidth]{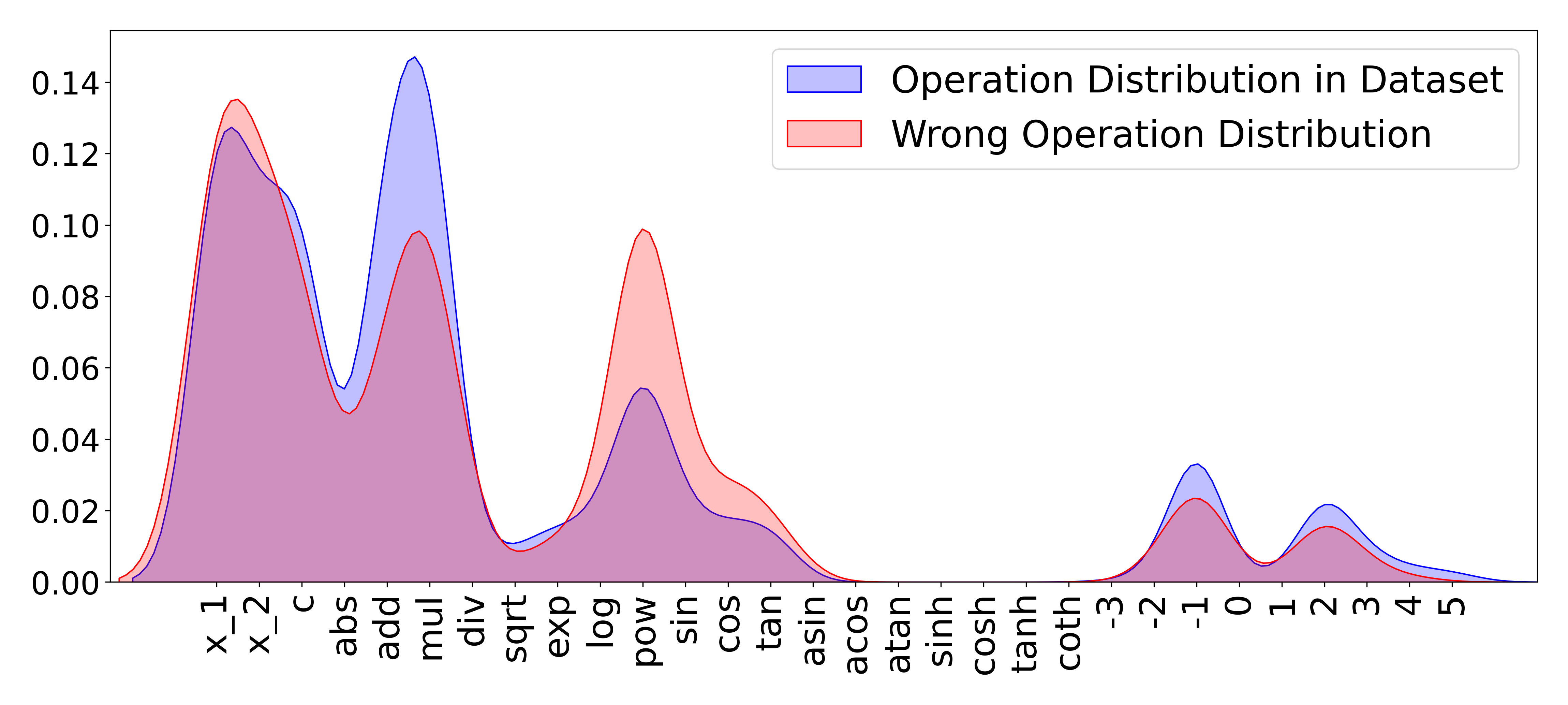}
\caption{Comparative analysis of operation frequency and error incidence. This graph contrasts the frequency of operations in the training set (blue) with the incidence of incorrect actions (red) observed in the SSDNC dataset. The y-axis represents error rates, while the x-axis lists the various operations. A notable correlation between these distributions is observed, suggesting that the agent tends to favor operations that it encountered more frequently during training.}
\label{dis-2}
\end{figure}

\noindent\textbf{Error rate trend throughout the expression generation process.} In addition to overall step accuracy, it is crucial to identify where the agent tends to make errors to thoroughly evaluate its behavior. To account for expressions of varying lengths, we normalized them by calculating their percentage of completion. Notably, the highest error rates occur during the middle stages of the process, as illustrated in \cref{depth}. A possible explanation for this trend is that initial decisions often involve straightforward binary operations such as addition ($+$) and multiplication ($\times$), which typically follow a general strategy of expanding the expression tree early on. Conversely, decisions made towards the end of the process tend to be easier as the expression becomes more focused and specific.

\noindent\textbf{Distribution of incorrect selections.} Our analysis indicates that the frequency of errors closely aligns with the distribution of operations within the training dataset. Specifically, the agent tends to favor operations that it encountered more frequently during training, as clearly depicted in \cref{dis-2}. This observation suggests a potential imbalance in the design of the training dataset, which should be carefully considered and addressed in future work.

\begin{figure}[h!]
\centering
\includegraphics[width=\columnwidth]{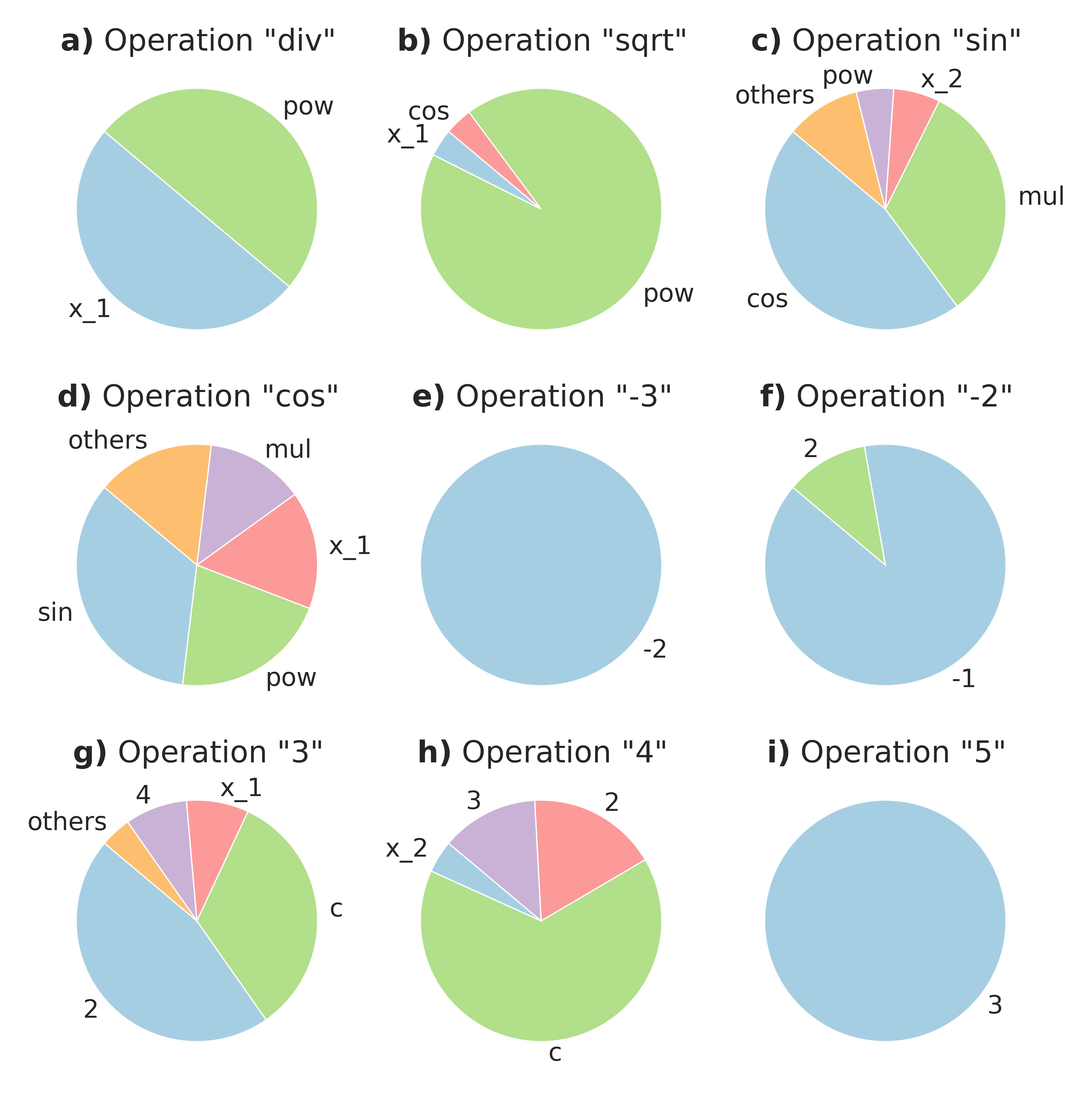}
\caption{Distribution of incorrect decision choices. The pie charts illustrate the decisions made during instances of incorrect decision-making. Each segment of the chart corresponds to an incorrect choice in relation to a specific correct operation. The size of each segment represents the proportion of these incorrect choices, summing to a total of $1$. The title above each chart identifies the correct operation targeted in each specific instance.}
\label{fig: Wrong choice distributions}
\end{figure}

\noindent\textbf{Analysis of specific error types.} Our observations indicate that the agent often struggles with determining exact constant values. While it correctly identifies the need to include a constant, the chosen value is often  incorrect, as illustrated in \cref{fig: Wrong choice distributions} \textbf{e)} to \textbf{i)}. This issue arises because the model does not perform coefficient fitting for constants, instead treating each discrete value as a separate action. Additionally, the agent commonly confuses $\sin(\cdot)$ with $\cos(\cdot)$ functions, likely due to their similar mathematical properties, as illustrated in \cref{fig: Wrong choice distributions} \textbf{c)} and \textbf{d)}. Another notable observation is the frequent confusion between the square root operation ($\sqrt{\cdot}$) and the power operation, as illustrated in \cref{fig: Wrong choice distributions} \textbf{g)}. This confusion is understandable, given that the square root is equivalent to raising a number to the power of $\frac{1}{2}$. Similarly,  the division operation ($\div$) is often mistaken for the power operation, as division is mathematically equivalent to raising a number to the power of $-1$, as shown in \cref{fig: Wrong choice distributions} \textbf{a)}. These patterns suggest that the agent understands the effects of these operations on mathematical relationships.

\section{Sym-Q model}
\label{sec: sym-q model}
\subsection{Overview}
\label{sec:method}
We propose Sym-Q, an RL-based method designed to tackle large-scale symbolic regression problems, featuring a collaborative co-design mechanism (see \cref{fig: framework}).  Leveraging our proposed offline RL training approach, Sym-Q facilitates interactive symbolic regression that integrates iteratively domain expert knowledge. As illustrated in \cref{fig: framework}.\textbf{b)}, Sym-Q effectively learns from both optimal demonstrations and sub-optimal reward signals, enhancing its adaptability and performance in complex regression tasks.

To establish a robust co-design symbolic regression framework, we conceptualize the generation of an expression tree as a Markov Decision Process (MDP). In this framework, the agent is trained to interpret observations and sequentially expand and refine the expression tree (\cref{fig: framework}.\textbf{b}). The \textit{state} is defined by the current expression tree and the observed data at each step, while actions involve selecting mathematical expressions to populate the nodes of the expression tree (\cref{fig: framework}.\textbf{c}). The effectiveness of these actions is quantified using the coefficient of determination $(R^2)$~\cite{glantz2001primer} as the reward metric, providing a clear measure of model accuracy. The agent sequentially replaces placeholders and expands the expression tree until the process concludes. Expression trees are generated using a beam search, repeated $128$ times as in prior works~\cite{biggio2021neural,li2022transformer}, resulting in $128$ distinct skeletons. The final expression is derived by applying the Broyden–Fletcher–Goldfarb–Shanno (BFGS)~\cite{fletcher2000practical} optimization on the constant terms in the expressions. The expression with the lowest fitting error is selected as the agent's final answer.

This formulation as a sequential decision process enriches the co-design mechanism by enabling domain experts to inject explicit prior knowledge based on the problem's context. Such expertise guides the model's decisions, allowing for dynamic refinement and optimization of the expression tree. Moreover, operations can be adjusted or corrected at any step based on current expert insights, ensuring that the expressions generated are both precise and pertinent.

\subsection{Expression tree}
\label{subsec: expression tree}
An expression tree, as defined by \citet{hopcroft2006automata}, is a hierarchical structure where internal nodes represent mathematical operations, such as addition, subtraction, and logarithm, while leaf nodes denote  constants or variables. This tree-based representation is particularly advantageous for sequential decision-making processes, as it allows for the incremental construction and refinement of mathematical expressions. Illustrated in \cref{fig: framework}.\textbf{a)}, the expression tree facilitates the systematic assembly and analysis of complex mathematical expressions, especially when their exact forms are initially unknown.

\subsection{Symbolic regression as a sequential decision-making task} \label{subsec: symbolic regression as a sequential decision-making task}
Symbolic regression involves searching for a sequence of operations $a_{1:T} = a_1, a_2, ..., a_T$ to construct an expression tree that accurately fits observational data. In this context, each operation $a_t$ represents a mathematical function either unary or binary operation within a predefined search space at decision step $t$. Previous large-scale transformer-based models~\cite{biggio2021neural,li2022transformer} address symbolic regression similarly to machine translation. These models use feature extractors to process input data points and utilize a transformer to decode these features into expression skeletons. In this setup, the model’s next-token predictions during training are conditioned on ground-truth tokens, resulting in exposure bias. This approach  causes the model to become overly reliant on teacher-forcing supervision, creating a significant discrepancy between training and inference phases.  As a result,   the model often struggles during real-world predictions, where ground-truth tokens are unavailable, leading to error accumulation.

In contrast, our approach utilizes offline reinforcement learning, eliminating the reliance  for teacher-forcing supervision at the decoder level. Instead, the model iteratively uses its own predictions as inputs during decoding, ensuring that the training process  closely align with real inference conditions. This approach also prevents direct gradient backpropagation across multiple tokens, promoting  more stable and realistic model updates\cite{bahdanau2016actor,xu2019rethinking,tsai2020order}. Our study conceptualizes symbolic regression as a Markov decision process (MDP). An MDP is characterized by the tuple $(S,A,r,P,\rho)$, where $S$ denotes the set of states reflecting the current situation, $A$ represents the set of possible actions, $r(s,a)$ is the reward function, $P(s'|s,a)$ defines the state transition probabilities, and $\rho(s)$ is the initial state distribution.

At each decision step $t$, the agent selects an action $a_t$ based on the policy $\pi_{\theta}(a_t|s_t)$, where $s_t$ represents the current state. This state encompasses both the observed data points $({X},{y})$ and the encoded structure of the current expression tree, derived from previous actions $a_{1:t-1}$. The action $a_t$  determines the next operation in the sequence, thereby incrementally building the symbolic expression $a_{1:T} = a_1, a_2, ..., a_T$ as a trajectory of the MDP, denoted by $\tau={s_1, a_1, s_2, a_2, ..., s_T, a_T}$. In our model, rewards are assigned exclusively at the termination state, with all intermediate states receiving a reward of $0$. We utilize the coefficient of determination $R^2$ as the reward metric, providing a clear measure of model accuracy. Under this MDP formulation, the objective of the RL-based agent is to discover a policy $\pi$ that maximizes the cumulative reward $J(\pi)$:

\begin{equation}
  J(\pi) = \mathbb{E}_{\tau \sim \rho_{\pi}}{\sum_{t=0}^{\infty} r(s_t, a_t)} ~.
\label{eq:j}
\end{equation}

Symbolic regression targets only optimal solutions, with all sub-optimal solutions receiving significantly lower rewards. To address the challenges of sparse reward and a large search space inherent in symbolic regression, we employ offline RL, a learning paradigm that merges the ability of supervised learning to leverage existing data with RL's capacity to optimize arbitrary rewards and leverage temporal compositionality~\cite{levine2020offline, kostrikov2021offline, kumar2020conservative, janner2021offline}.

Under the offline RL framework, the objective remains the optimization of the function specified in \cref{eq:j}, However, because the agent cannot interact directly with the environment, it must adopt a different approach to learn from the static dataset. By relying on a predetermined, static dataset of transitions denoted as $D = {(s^i_t, a^i_t, s^i_{t+1}, r^i_t)}$, offline RL circumvents the inefficiencies of online exploration and mitigates the impact of sparse reward signals. By leveraging this approach, Sym-Q efficiently utilizes existing data to train the agent, overcoming the limitations of online exploration and reward sparsity. 
\subsection{Symbolic Q-network} \label{subsec: symbolic q-network}

Sym-Q is an offline RL algorithm specifically designed to address large-scale symbolic regression problems by enabling sequential decision-making and efficient step-wise updates. Sym-Q comprises three integral modules:

\begin{enumerate}
    \item Point set encoder $E^{points}_{\phi}({{X,y}})$ which transforms the point sets ${(X,y)}=((x_1^1,x_1^2,y_1),(x^1_2,x_2^2,y_2), ..., (x^1_n,x_n^2,y_n))$ associated with each equation into a latent space, resulting in a fixed-size latent representation $z^p \in \mathbb{R}^{1\times K_p}$, where $K_p$ denotes  the dimensionality of the latent variable; 
    \item Expression tree encoder $E^{tree}_{\psi}(M_t)$ which maps the current tree's one-hot encoded matrix $M_t$ into another fixed-size latent representation $z^t \in \mathbb{R}^{1\times K_t}$, where $K_t$ indicates the dimensionality of the variable; 
    \item  Q-network $Q_{\theta}(s_t)$ which calculates the Q-value for each potential operation, given the combined latent state $s_t=(z^p,z^t)$. 
\end{enumerate}

This architecture enables Sym-Q to address symbolic regression in a novel and efficient way. The comprehensive framework of Sym-Q is depicted in \cref{fig: framework}. For further details on the neural network architectures and hyperparameters used, please refer to \cref{appendix: details of neural networks and hyperparameters}. Additionally, we conducted experiments using various combinations of encoders, including Transformer and RNN architectures, paired with different loss functions such as cross entropy and Classical mean squared error for Q-Learning. Our findings demonstrate that Sym-Q is the first large-scale symbolic regression model capable of processing sequence data without depending on a transformer-based architecture. This novel approach not only offers a more lightweight and modular framework but also underscores the potential of reinforcement learning-based methods for tackling complex symbolic regression tasks. For further details, refer to \cref{appendix: RNN and loss}
\subsection{Efficient conservative offline Q-learning for symbolic regression}
\label{subsec: efficient conservative offline Q-learning for symbolic regression}
In training the Sym-Q through offline RL, a significant challenge is the overestimation of values due to distributional shifts between the dataset and the learned policy. To address this, we adopt Conservative Q-learning (CQL)~\cite{kumar2020conservative}, which minimizes the values of state-action pairs outside the training data distribution, while simultaneously maximizing the values within this distribution. Given the sparse reward structure of symbolic regression, where rewards are typically given only at the completion of each trajectory and most sub-optimal trajectories receive lower rewards, we have adapted CQL to this context. Our modified version of CQL is customized for symbolic regression, aiming to effectively leverage optimal offline data in this domain:

\begin{equation}
J(\theta,\psi) = -\mathbb{E}_{(s,a^i)\sim \mathcal{D}}\left[\log\left(\frac{e^{Q_{\theta,\psi}(s,a^i)}}{\sum_{j=1}^m e^{Q_{\theta,\psi}(s,a^j)}}\right)\right] ,
\label{CQL-SR}
\end{equation}

where $Q_{\theta,\psi}$ represents the network and its weight parameters, including those beyond the points encoder. $D$ represents the training dataset, $a^i$ is the action demonstrated at state $s$, and ${a^j | j \neq i, 1 \leq j \leq m}$ are the non-demonstrated actions, regarded as suboptimal.

Building on the relationship between conservative Q-learning loss and cross-entropy loss, our proposed objective function represents the log probability of the optimal action. This mirrors the emphasis of cross-entropy on identifying the correct class. The inclusion of the softmax function in the denominator serves to normalize the Q values across all possible actions, effectively converting them into a probability distribution.

By adopting this objective function, our approach more closely aligns with the principles of traditional supervised learning. This alignment makes the method more intuitive and straightforward to implement, particularly in scenarios where the correct action at each state is well-defined, such as in symbolic regression.

\subsection{Supervised contrastive learning for points encoder}
\label{subsec: supervised contrastive learning for points encoder}
Following a similar idea from \citet{li2022transformer,dong2023SimMMDG}, we implemented the supervised contrastive loss for expressions and data points that share the same skeleton. This approach uses the skeleton of expressions as category labels to enrich supervisory information. More specifically, expressions and their corresponding latent points encoding $z^p_i$, which belong to the same skeleton, are grouped together in the embedding space. At the same time, we ensure that the latent points encoding $z^p_j$ from different skeletons are distanced from each other. The supervised contrastive loss is defined as:

\begin{equation}
\begin{aligned}
 J(\phi) &= \sum_{i=1}^{N} \frac{-1}{|\mathcal{P}(i)|} \sum_{p \in \mathcal{P}(i)} \log
 \frac{\exp(E^{points}_{\phi}({X,y})^i) \cdot E^{points}_{\phi}({X,y})^p / \tau)}{\sum_{j=1}^{N} 1_{[j \neq i]} \exp(E^{points}_{\phi}({X,y})^i \cdot E^{points}_{\phi}({X,y})^j / \tau)}
\end{aligned}
\end{equation}

where $\mathcal{P}(i)$ represents the set of indices for all positives in the multiviewed batch
distinct from $i$, and $|\mathcal{P}(i)|$ is its cardinality; $\tau$ is an adjustable scalar temperature parameter controlling class separation; $N$ represents the mini-batch size; $z^p_i=E^{points}_{\phi}({X,y})^i$ is the points embedding of the sample $i$ in a batch, $z^p_p=E^{points}_{\phi}({X,y})^p$ is the points embedding of a positive sample sharing the same skeleton as sample $i$. The overall loss is then given by:

\begin{equation}
    \begin{aligned}
     J(\phi,\theta,\psi) = \alpha_1 J(\theta,\psi) +\alpha_2 J(\phi) ,
    \end{aligned}
    \label{offline RL}
\end{equation}

where $\alpha_{1,2}$ are scalar weighting hyperparameters. More details about the parameters settings and network architectures can be found in \cref{appendix: details of neural networks and hyperparameters}.

\subsection{Co-design mechanism}
\label{subsec: co-design}
A central feature of the proposed architecture is the innovative co-design mechanism, which integrates users' domain knowledge with the learned knowledge of the model, allowing for interaction at any decision step.

This collaborative approach leverages the computational power and pattern recognition capabilities of a reinforcement learning agent while incorporating the contextual knowledge and insights from human experts. By involving domain experts, the generated mathematical expressions become more accurate, relevant and meaningful. This synergy between human expertise and machine learning ensures that the model is grounded in the true underlying physical processes and dynamics. Consequently, the collaboration facilitates more effective problem-solving, leading to impactful and meaningful solutions. This dynamic interplay between human and machine intelligence is crucial for addressing the limitations of purely automated methods, fostering a deeper understanding of complex data relationships and enhancing the model's adaptability to diverse and evolving problem contexts.

This co-design approach can be conceptualized as an environment within a reinforcement learning framework. At each interaction step, the agent accesses both the point sets and the current expression tree. The expression tree, or parts of it, can originate from three potential sources: 1) decisions made by Sym-Q, 2) hypotheses or modifications introduced through direct human intervention, and 3) partial decision sequences suggested by other symbolic regression models. In this work, we focus only on the first two types of decisions as prior information to the Sym-Q agent. The co-design mechanism enables domain experts to integrate explicit prior knowledge tailored to the problem's context. Their expertise guides the model's decisions, enabling dynamic refinement and optimization of the expression tree. Additionally, operations can be modified or corrected at any stage based on domain expert insights, ensuring that the generated expressions are both accurate and relevant.

\begin{figure}[th!]
    \centering
    \includegraphics[width=\columnwidth]{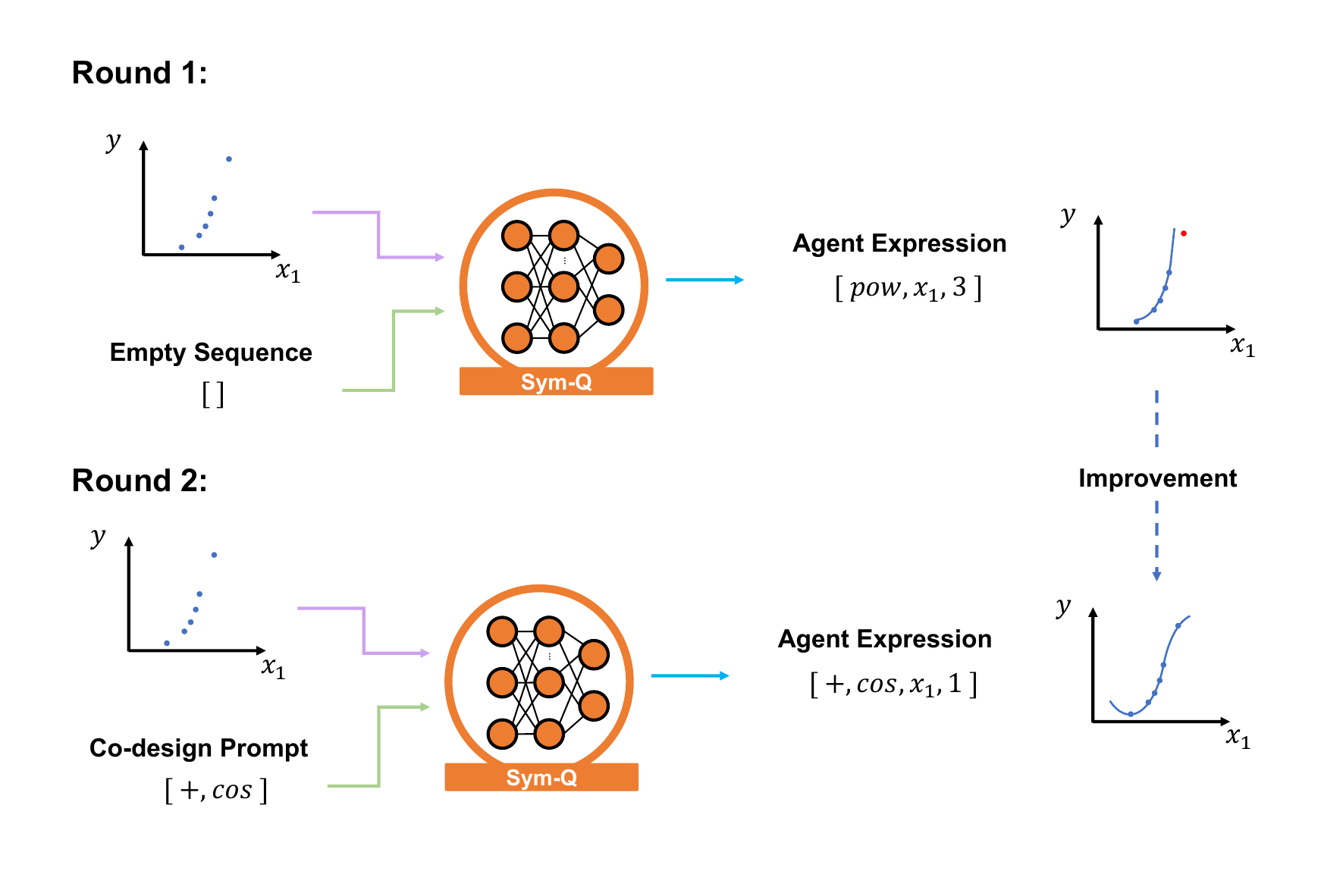}
    \caption{Iterative Enhancement through Co-Design in Sym-Q.
    This diagram illustrates the iterative refinement process within Sym-Q when initial solutions are sub-optimal.}
    \label{fig: co-design process}
\end{figure}

 Our proposed co-design process illustrated  in \cref{fig: co-design process}, begins with the agent attempting to construct an expression starting from an empty sequence. This initial attempt often results in an expression that only partially fits  the data points, revealing a misalignment between the agent's interpretation  of the data structure and  an accurate solution. In the next  step, we employ a co-design mechanism to guide the agent toward a more accurate representation. To expand the expression tree, we introduce a binary operator as the initial token, followed by a specific operator that better captures the shape of the data points. This step leverages human intuition to correct the agent’s initial misjudgments, enhancing its problem-solving capabilities. By integrating insights into the agent's behavior with domain expertise, our co-design process becomes both realistic and reliable. This integration  significantly improves the agent's ability to generate accurate expressions, making it more effective in addressing complex problems.

\section{Discussion}
\label{sec: conclusion}

 The study presents a novel paradigm for large-scale symbolic regression  by explicitly framing  it as a sequential decision-making task and addressing it with the proposed Sym-Q algorithm. Sym-Q not only excels at discovering symbolic expressions from scratch but also overcomes a key limitation of traditional symbolic regression models: their less effectiveness and efficiency in incorporating human priors due to limited generalization and extrapolation capabilities, particularly when faced with fixed or out-of-distribution (OOD) equation structures.
 
 To address this challenge, Sym-Q integrates a co-design mechanism that enables more effective interactive refinement  of expressions. This innovative approach allows domain experts to provide partially defined expression trees, fostering real-time collaboration with the model. By dynamically modifying generated nodes or providing prior information, experts can more effectively guide the agent to develop  mathematical expressions that accurately capture  underlying problem dynamics and adhere to established physical laws, particularly when partial domain knowledge is available. This co-design mechanism effectively integrates expert insights and  user hypotheses, enhancing both the interpretability and scientific relevance of the resulting expressions.


A key strength of Sym-Q lies in its versatility, as it can be combined with various types of encoders, from advanced transformer architectures to simpler recurrent neural networks, for processing expression trees. By leveraging reinforcement learning, Sym-Q guides the model step-by-step through the equation construction process, enabling the agent to learn how to build expressions from observations rather than merely memorizing patterns.  Notably, Sym-Q is the first approach to successfully tackle the large-scale symbolic regression task without relying on a transformer decoder. Extensive benchmark evaluations demonstrate  that Sym-Q outperforms  other state-of-the-art algorithms in skeleton recovery rate while also achieving superior fitting accuracy and recovery rates across the majority of benchmark datasets used for large-scale symbolic regression models.

This innovative framework is particularly well-suited  for co-design, as evidenced   by the  demonstrated  effectiveness of Sym-Q's co-design mechanism in the evaluation experiments. By integrating domain knowledge, Sym-Q effectively tackles complex scenarios, such as recovering drift terms in the Feynman dataset and deriving analytical expressions from synthetic transit spectra. This integration leads to notable improvements in both recovery rates and the coefficient of determination. The experiments illustrate that as more domain expertise is incorporated, the co-design approach not only enhances performance metrics but also aligns the model's outputs with underlying physical laws and patterns. Compared to NeSymReS, Sym-Q’s co-design mechanism consistently delivers superior results, showcasing more reliable improvements across varying levels of ground truth availability and  effectively leveraging partial information.

Sym-Q’s co-design capabilities unlock significant advancements in symbolic regression, particularly in scenarios involving out-of-distribution (OOD) equation structures.  By effectively leveraging partial prior knowledge and dynamically adapting to unfamiliar equation forms, Sym-Q demonstrates robust generalization and adaptability, making it a powerful tool for tackling complex, real-world problems.


A promising direction for future research involves extending Sym-Q to tackle more complex types of expressions, such as Ordinary Differential Equations (ODE) and Partial Differential Equations (PDE). Additionally, further enhancements to the co-design framework could explore more sophisticated ways of integrating expert knowledge, potentially incorporating comprehensive constraints or hierarchical guidance to further refine the discovery process.

\backmatter

\bmhead{Acknowledgements}

The contributions of Yuan Tian, were funded by the ETH grant  no. ETH-12 21-1. The contributions of Wenqi Zhou and Michele Viscione were funded by the Hasler Foundation no. 24019.

\section*{Declarations}

We declare that none of the results have been published or are under consideration elsewhere. 


\noindent






\clearpage
\bibliography{sn-bibliography}

\clearpage
\appendix
\renewcommand{\thetable}{\Alph{section}\arabic{table}}

\section{Details of Neural Networks and Hyperparameters}
\label{appendix: details of neural networks and hyperparameters}
The Sym-Q consists of four main components: the points encoder, the tree encoder, the fusion layer, and the Q-network. For the points encoder, we apply the Set Transformer architecture~\cite{devlin2018bert,biggio2021neural}, using the original publicly available implementation, which has been proven effective for large-scale symbolic regression tasks~\cite{biggio2021neural,bendinelli2023controllable}. According to the original paper's notation, our encoder is composed of multiple Induced Set Attention Blocks (ISABs) and a final component performing Pooling by Multihead Attention (PMA). This design reduces the computational burden associated with the self-attention operation and aggregates the encoder's output into trainable abstract features, providing a compressed representation of the input data points. For the tree encoder, we employ a standard Transformer encoder~\cite{vaswani2017attention}, using the default PyTorch implementation. In our research,  the fusion layer is a fully connected layer with 4096 neurons. Similarly, the Q-network consists of an MLP with four fully connected hidden layers, each containing 4096 neurons, responsible for predicting the Q-value. The GELU activation function~\cite{hendrycks2016gaussian} is applied in all hidden layers of both the fusion layer and Q-network.  Additional hyper-parameters can be found in \autoref{tab:hyperparameters}. All the experiments mentioned in this paper were conducted using a single NVIDIA GeForce RTX 3090 GPU.

\begin{table}[h!]
    \centering
    \begin{tabular*}{0.8\textwidth}{@{\extracolsep\fill}lc}
        \toprule
        \multicolumn{2}{c}{\textbf{Points encoder}} \\
        \midrule
        Number of ISABs & 5 \\
        Hidden dimension &  512 \\
        Number of heads & 8 \\
        Number of PMA features & 100 \\
        Number of inducing points & 100 \\
        \midrule
        \multicolumn{2}{c}{\textbf{Tree encoder}} \\
        \midrule
        Number of layers &  5 \\
        Hidden dimension &  512 \\
        Number of heads & 8 \\
        \midrule
        \multicolumn{2}{c}{\textbf{Training}} \\
        \midrule
        Batch size & 512 \\
        Learning rate & 0.0001 \\
        Scale weight \(\alpha\) & 0.2 \\
        Temperature parameter $\tau$ & 0.07 \\
        Max norm of gradients & 1.0 \\
        \midrule
        \multicolumn{2}{c}{\textbf{Inference}} \\
        \midrule
        Beam size & 128 \\
        Restart times of BFGS & 20 \\
        \bottomrule
    \end{tabular*}
    \caption{Hyperparameters for our models.}
    \label{tab:hyperparameters}
\end{table}

\clearpage
\section{Details of Dataset Generation and Action Space}
\label{appendix: details of dataset generation and action space}
The training set and SSDNC benchmark are generated under the guidelines provided by \citet{li2022transformer}. We sample each non-leaf node following the unnormalized distribution shown in \cref{tab:unormalized probabilities of unary and binary operators used by the dataset generator}. The database approximately contains $100,000$ unique expression skeletons. Then the training set is generated by sampling $50$ numerical constants for each skeleton. The SSDNC benchmark includes approximately $100$ unique expression skeletons and $10$ sampled numerical constants for each skeleton. 

\begin{table}[h!]
    \begin{tabular*}{0.8\textwidth}{@{\extracolsep\fill}lcc}
        \toprule
        \textbf{Operation} & \textbf{Mathematical meaning} & \textbf{Unnormalized probability} \\
        \midrule
        add & $+$ & 10 \\
        mul & $\times$ & 10 \\
        sub & $-$ & 5 \\
        div & $\div$ & 5 \\
        sqrt & $\sqrt{}$ & 4 \\
        exp & $\exp$ & 4 \\
        ln & $\ln$ & 4 \\
        sin & $\sin$ & 4 \\
        cos & $\cos$ & 4 \\
        tan & $\tan$ & 4 \\
        pow2 & $(\cdot)^2$ & 4 \\
        pow3 & $(\cdot)^3$ & 4 \\
        pow4 & $(\cdot)^4$ & 2 \\
        pow5 & $(\cdot)^5$ & 1 \\
        \botrule
    \end{tabular*}
    \caption{Unnormalized probabilities of unary and binary operators used by the dataset generator.}
    \label{tab:unormalized probabilities of unary and binary operators used by the dataset generator}
\end{table}

In this work, at every decision step $t$, we consider $30$ different operations, which is listed below in \cref{tab:action space and corresponding id}.

\begin{table}[h!]
    \begin{tabular*}{0.6\textwidth}{@{\extracolsep\fill}lc}
        \toprule
        \textbf{Operation} & \textbf{Operation id} \\
        \midrule
        $x_1$ & $0$ \\
        $x_2$ & $1$ \\
        $c$ & $2$ \\
        $\text{abs}$ & $3$ \\
        $+$ & $4$ \\
        $*$ & $5$ \\
        $/$ & $6$ \\
        $\sqrt{}$ & $7$ \\
        $\exp$ & $8$ \\
        $\log$ & $9$ \\
        $**$ & $10$ \\
        $\sin$ & $11$ \\
        $\cos$ & $12$ \\
        $\tan$ & $13$ \\
        $\sin^{-1}$ & $14$ \\
        $\cos^{-1}$ & $15$ \\
        $\tan^{-1}$ & $16$ \\
        $\sinh$ & $17$ \\
        $\cosh$ & $18$ \\
        $\tanh$ & $19$ \\
        $\coth$ & $20$ \\
        $-3$ & $21$ \\
        $-2$ & $22$ \\
        $-1$ & $23$ \\
        $0$ & $24$ \\
        $1$ & $25$ \\
        $2$ & $26$ \\
        $3$ & $27$ \\
        $4$ & $28$ \\
        $5$ & $29$ \\
        \botrule
    \end{tabular*}
    \caption{Action space and corresponding id.}
    \label{tab:action space and corresponding id}
\end{table}

\clearpage
\section{Fitting Accuracy on Additional Benchmarks}
\label{appendix: fitting accuracy on additional benchmarks}

We evaluate our method and current state-of-the-art approaches on the widely used public benchmarks, including Nguyen benchmark~\cite{uy2011semantically}, the Constant benchmark, the Keijzer benchmark~\cite{keijzer2003improving}, R rationals~\cite{krawiec2013approximating}, the AI-Feynman dataset~\cite{udrescu2020ai}, as well as the SSDNC dataset~\cite{petersen2019deep}. Nguyen was the main benchmark used in~\citet{petersen2019deep}. In our study, we consider all the expressions with up to two variables. From the results in \cref{tab:benchmarks}, our method outperforms all baseline methods in terms of weighted average $R^2$ on six benchmarks. The weighted average $R^2$ is calculated based on the number of expressions in each benchmark. In \cref{tab:Nguyen} - \cref{tab:AI-Feynman}, we show the name of the benchmark and corresponding expressions.

\begin{table}[h!]
    \centering
    \resizebox{\textwidth}{!}{%
        \begin{tabular}{@{\extracolsep\fill}lccccccc}
            \toprule
            \textbf{Benchmark} & \textbf{Number of expressions} & \textbf{Sym-Q}  & \textbf{T-JSL} & \textbf{NeSymReS} & \textbf{SymbolicGPT} & \textbf{DSR}  & \textbf{GP}  \\
            \toprule
            Metric&  & $R^2$ & $R^2$ &  $R^2$ &  $R^2$ & $R^2$ & $R^2$ \\
            \midrule
            Nguyen & 12 & 0.86482 & \textbf{0.99999} & 0.97538 & 0.64394 & 0.99489 & 0.89019 \\
            Constant & 8 & 0.86860 & \textbf{0.99998} & 0.84935 & 0.69433 & 0.99927 & 0.90842 \\
            Keijzer & 12 & 0.94392 & 0.98320 & 0.97500 & 0.59457 & 0.96928 & 0.90082 \\
            R & 3 & \textbf{0.99999} & \textbf{0.99999} & 0.84935 & 0.71093 & 0.97298 & 0.83198 \\
            AI-Feynman & 15 & 0.99975 & \textbf{0.99999} & \textbf{0.99999} & 0.64682 & \textbf{0.99999} & 0.92242 \\
            SSDNC & 963 & \textbf{0.95135} & 0.94782 & 0.85792 & 0.74585 & 0.93198 & 0.88913 \\
            \midrule
            Weighted avg. & 1013 & \textbf{0.95044} & 0.95020 & 0.86271 & 0.74087 & 0.93483 & 0.88976 \\
            \botrule
        \end{tabular}
    }
    \caption{Results comparing our Sym-Q with state-of-the-art methods on several benchmarks. All methods use the beam search strategy with the beam size equaling 128. We report the average value of $R^2$ for each benchmark and the weighted average $R^2$ on six benchmarks.}
    \label{tab:benchmarks}
\end{table}

\begin{table}[h!]
    \begin{tabular*}{\textwidth}{@{\extracolsep\fill}lc}
    \toprule
        \textbf{Name} & \textbf{Expression} \\
        \midrule
        Nguyen-1 & $x^3+x^2+x$ \\
        Nguyen-2 & $x^4+x^3+x^2+x$ \\
        Nguyen-3 & $x^5+x^4+x^3+x^2+x$ \\
        Nguyen-4 & $x^6+x^5+x^4+x^3+x^2+x$ \\
        Nguyen-5 & $\sin(x^2)\cos(x)-1$ \\
        Nguyen-6 & $\sin(x)+\sin(x+x^2)$ \\
        Nguyen-7 & $\ln(x)+\ln(x^2+1)$ \\
        Nguyen-8 & $\sqrt{x}$ \\
        Nguyen-9 & $\sin(x)+\sin(y)$ \\
        Nguyen-10 & $2\sin(x)\cos(y)$ \\
        Nguyen-11 & $x^y$ \\
        Nguyen-12 & $x^4-x^3+\frac{1}{2}y^2-y$ \\
        \botrule
    \end{tabular*}
    \caption{Benchmark functions in Nguyen.}
    \label{tab:Nguyen}
\end{table}

\begin{table}[h!]
    \begin{tabular*}{\textwidth}{@{\extracolsep\fill}lc}
        \toprule
        \textbf{Name} & \textbf{Expression} \\
        \midrule
        Constant-1 & $3.39x^3 + 2.12x^2 + 1.78x$ \\
        Constant-2 & $\sin(x^2)\cos(x)-0.75$ \\
        Constant-3 & $\sin(1.5x)\cos(0.5y)$ \\
        Constant-4 & $2.7x^y$ \\
        Constant-5 & $\sqrt{1.23x}$ \\
        Constant-6 & $x^{0.423}$ \\
        Constant-7 & $2\sin(1.3x)\cos(y)$ \\
        Constant-8 & $\ln(x + 1.4) + \ln(x^2 + 1.3)$ \\
        \botrule
    \end{tabular*}
    \caption{Benchmark functions in Constant.}
    \label{tab:Constant}
\end{table}

\begin{table}[h!]
    \begin{tabular*}{\textwidth}{@{\extracolsep\fill}lc}
        \toprule
        \textbf{Name} & \textbf{Expression} \\
        \midrule
        Keijzer-3 & $0.3x\sin(2\pi x)$ \\
        Keijzer-4 & $x^3 \exp(-x)\cos(x)\sin(x)(\sin(x^2)\cos(x)-1)$ \\
        Keijzer-6 & $\frac{x(x+1)}{2}$ \\
        Keijzer-7 & $\ln{x}$ \\
        Keijzer-8 & $\sqrt{x}$ \\
        Keijzer-9 & $\ln{x+\sqrt{x^2+1}}$ \\
        Keijzer-10 & $x^y$ \\
        Keijzer-11 & $xy+\sin{(x-1)(y-1)}$ \\
        Keijzer-12 & $x^4-x^3+\frac{y^2}{2}-y$ \\
        Keijzer-13 & $6\sin{x}\cos{y}$ \\
        Keijzer-14 & $\frac{8}{2+x^2+y^2}$ \\
        Keijzer-15 & $\frac{x^3}{5}+\frac{y^3}{2}-y-x$ \\
        \botrule
    \end{tabular*}
    \caption{Benchmark functions in Keijzer.}
    \label{tab:Keijzer}
\end{table}

\begin{table}[h!]
    \begin{tabular*}{\textwidth}{@{\extracolsep\fill}lc}
        \toprule
        \textbf{Name} & \textbf{Expression} \\
        \midrule
        R-1 & $\frac{(x+1)^3}{x^2-x+1}$ \\
        R-2 & $\frac{x^5-3x^3+1}{x^2+1}$ \\
        R-3 & $\frac{x^6+x^5}{x^4+x^3+x^2+x}$ \\
        \botrule
    \end{tabular*}
    \caption{Benchmark functions in R.}
    \label{tab:R}
\end{table}

\begin{table}[h!]
    \begin{tabular*}{\textwidth}{@{\extracolsep\fill}lc}
        \toprule
        \textbf{Name} & \textbf{Expression} \\
        \midrule
        Feynman-1 & $\frac{\exp{\frac{-x^2}{2}}}{\sqrt{2\pi}}$ \\
        Feynman-2 & $\frac{\exp{\frac{-(xy^{-1})^2}{2}}}{\sqrt{2\pi}y}$ \\
        Feynman-3 & $xy$ \\
        Feynman-4 & $xy$ \\
        Feynman-5 & $\frac{1}{2}xy^2$ \\
        Feynman-6 & $\frac{x}{y}$ \\
        Feynman-7 & $\frac{\sin{x}}{\sin{y}}$ \\
        Feynman-8 & $\frac{x}{y}$ \\
        Feynman-9 & $\frac{xy}{2\pi}$ \\
        Feynman-10 & $1.5xy$ \\
        Feynman-11 & $\frac{x}{4\pi y^2}$ \\
        Feynman-12 & $\frac{xy^2}{x}$ \\
        Feynman-13 & $xy^2$ \\
        Feynman-14 & $\frac{x}{2(1+y)}$ \\
        Feynman-15 & $\frac{xy}{2\pi}$ \\
        \botrule
    \end{tabular*}
    \caption{Benchmark functions in AI-Feynman.}
    \label{tab:AI-Feynman}
\end{table}

\clearpage
\section{Co-design Comparison}
\label{appendix: co-design comparison}

\begin{figure}[h]
\centering
\includegraphics[width=\columnwidth]{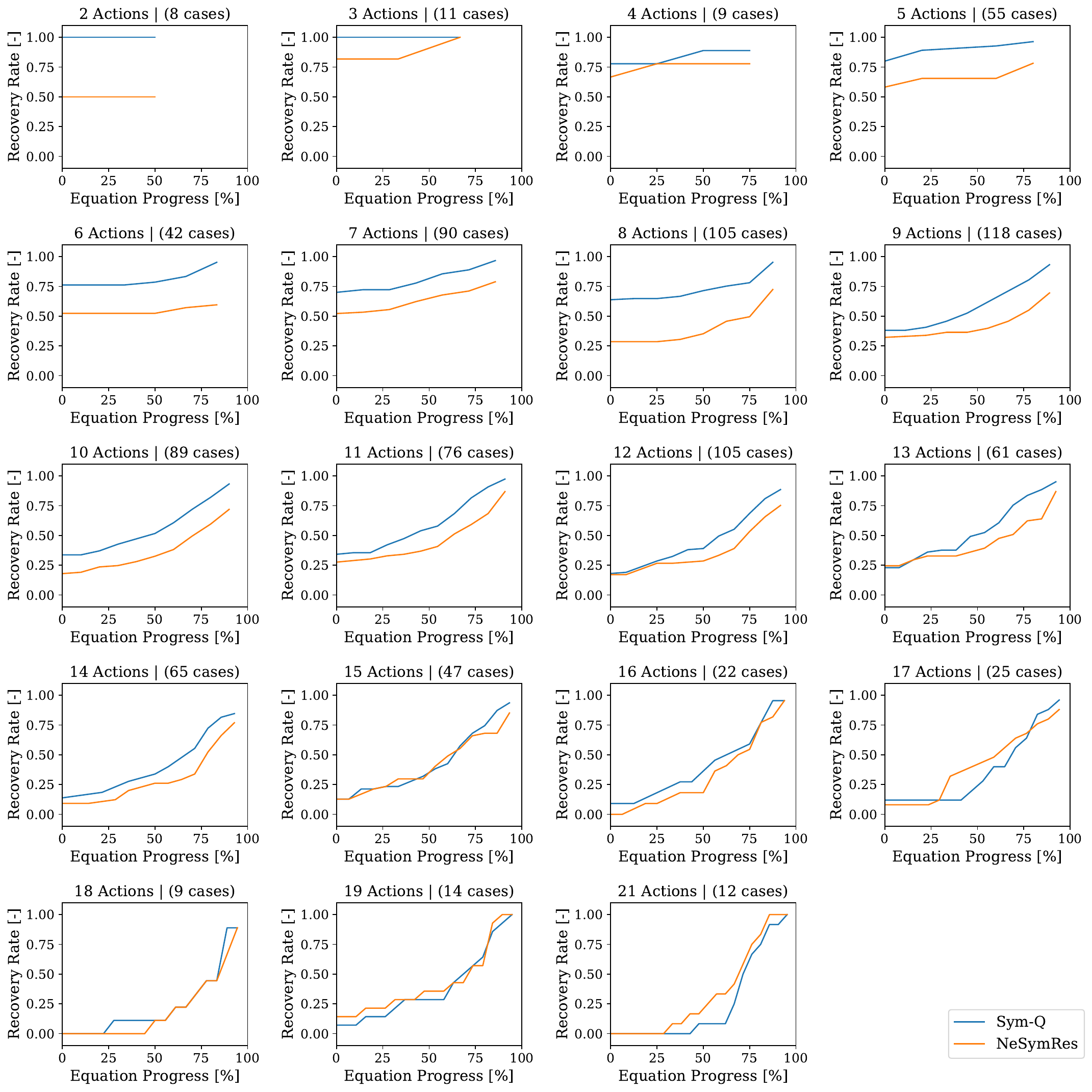}
\caption{Performance improvement comparison between Sym-Q and NeSymReS for different equation lengths, with each sub figure representing a distinct equation complexity (number of actions) and the corresponding number of cases. The recovery rate is plotted as a function of equation progress, expressed as a percentage of the total equation information provided. Experiment details are stated in \cref{fig: performance improvement via co-design}.}
\label{fig:co-design comparison}
\end{figure}

In addition to the experiment on performance improvement through the co-design of Sym-Q, we conducted a co-design evaluation on NeSymReS~\cite{biggio2021neural} to benchmark its effectiveness under similar conditions. To conduct the co-design for NeSymReS, we provide the known tokens to the decoder and let the model to predict the next token auto-regressively until it predicts the "end" token. The results are presented in \cref{fig:co-design comparison}. As illustrated, the recovery rate of Sym-Q consistently surpasses that of NeSymReS across various equation lengths, due to the superior initial performance of Sym-Q. This suggests that Sym-Q's co-design process offers a more robust foundation for recovery rate improvements.

\clearpage
\section{Assessment of Encoding Strategies}
\label{appendix: RNN and loss}
From an architectural perspective, a key distinction of Sym-Q lies in is its approach to generating expressions. Unlike auto-regressive transformer based models that rely on the attention mechanism of transformer decoders, Sym-Q utilizes the transformer solely as a sequence data encoder. This design choice makes the encoder modular, allowing it to be easily replaced with alternative sequence encoders, such as RNNs, without compromising functionality.

\begin{figure}[h]
\centering
\includegraphics[width=\columnwidth]{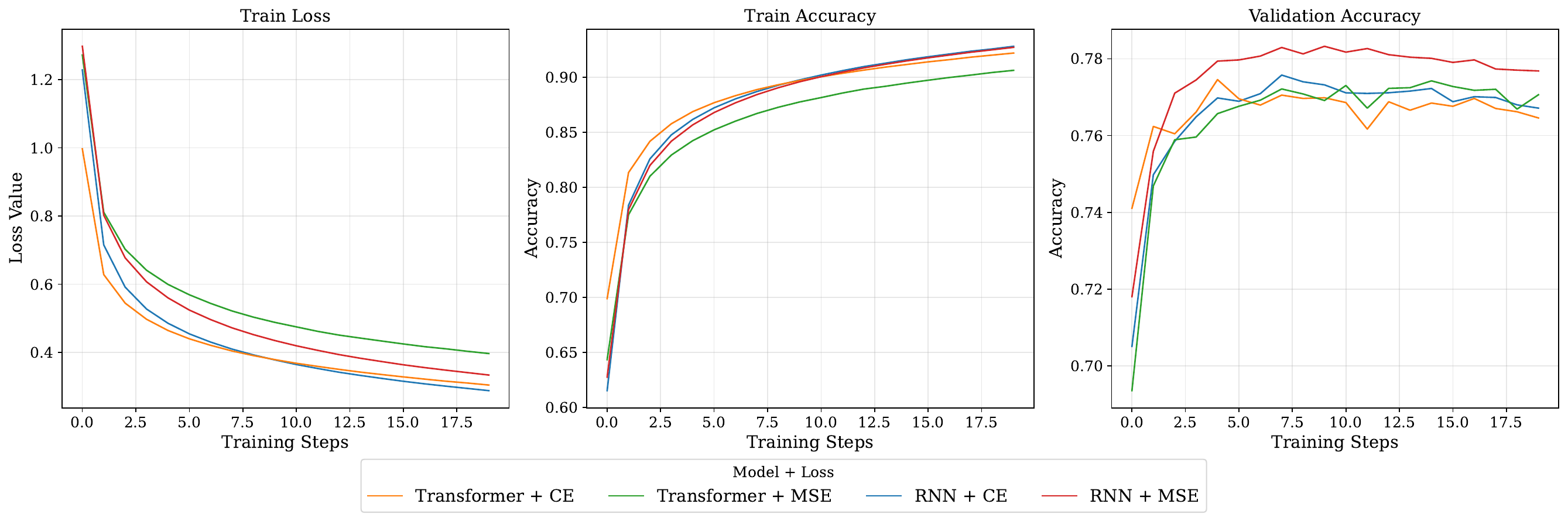}
\caption{Training and validation performance across different combinations of encoder (Transformer or RNN) and loss function (Cross Entropy or Mean Squared Error).}
\label{fig:interchangeable components}
\end{figure}

We conducted experiments with combinations of different encoders (Transformer or RNN) and loss functions (Cross Entropy or Mean Squared Error) on the same training and validation dataset. \cref{fig:interchangeable components} demonstrates that  Sym-Q’s performance remains roughly stable  across various  configurations, underscoring the flexibility and innovation of our framework. Notably, Sym-Q is the first large-scale symbolic regression model capable of handling sequence data without relying on a transformer architecture. This innovative approach not only results in a more lightweight and modular framework but also highlights the potential of reinforcement learning-based methods in addressing complex symbolic regression tasks.

\clearpage
\section{Ablation study on contrastive loss}
\label{appendix: Ablation study on contrastive loss}
We analyze different contrastive loss coefficients in this sensitivity analysis. Due to the limitation of computation cost, we conduct training sessions on a smaller dataset using three different values (0.1, 0.5, and 1.0) for the contrastive loss coefficient $\alpha_2$ with $\alpha_1$ fixed (loss function is defined in \cref{eq:j}). The results presented in \cref{tab:study on loss coefficient} indicate that model performance, as measured by the $R^2$ metric, remains stable across these different settings of $\alpha_2$. This consistency suggests that the framework's performance is relatively robust to changes in the contrastive loss coefficient.
\begin{table}[h!]
    \centering
        \begin{tabular*}{\textwidth}{@{\extracolsep\fill}lcccc}
            \toprule
            \textbf{Benchmark} & \textbf{Number of expressions} & \multicolumn{3}{c}{\textbf{Contrastive Loss}} \\
            \cmidrule(lr){3-5}
             & & $\alpha_2 = 0.1$ & $\alpha_2 = 0.5$ & $\alpha_2 = 1.0$ \\
            \midrule
            Nguyen     & 12 & 0.897040 & 0.922825 & 0.917503 \\
            Constant   & 8  & 0.814266 & 0.878594 & 0.826779 \\
            Keijzer    & 12 & 0.999564 & 0.858027 & 0.961467 \\
            R          & 3  & 0.996432 & 0.993851 & 0.998618  \\
            AI-Feynman & 12 &0.932763 & 0.957403 & 0.959037   \\
            \midrule
            Weighted avg. && 0.924592 & 0.912114 & 0.929068  \\
            \botrule
        \end{tabular*}
        \caption{The effect of varying the contrastive loss coefficient $\alpha_2$ on model performance across benchmarks. Model performance is evaluated using the $R^2$ metric.}
    \label{tab:study on loss coefficient}
\end{table}

\clearpage
\section{Evaluation on SRBench}
\label{appendix: evaluation on srbench}

\begin{figure}[h]
\centering
\includegraphics[width=\columnwidth]{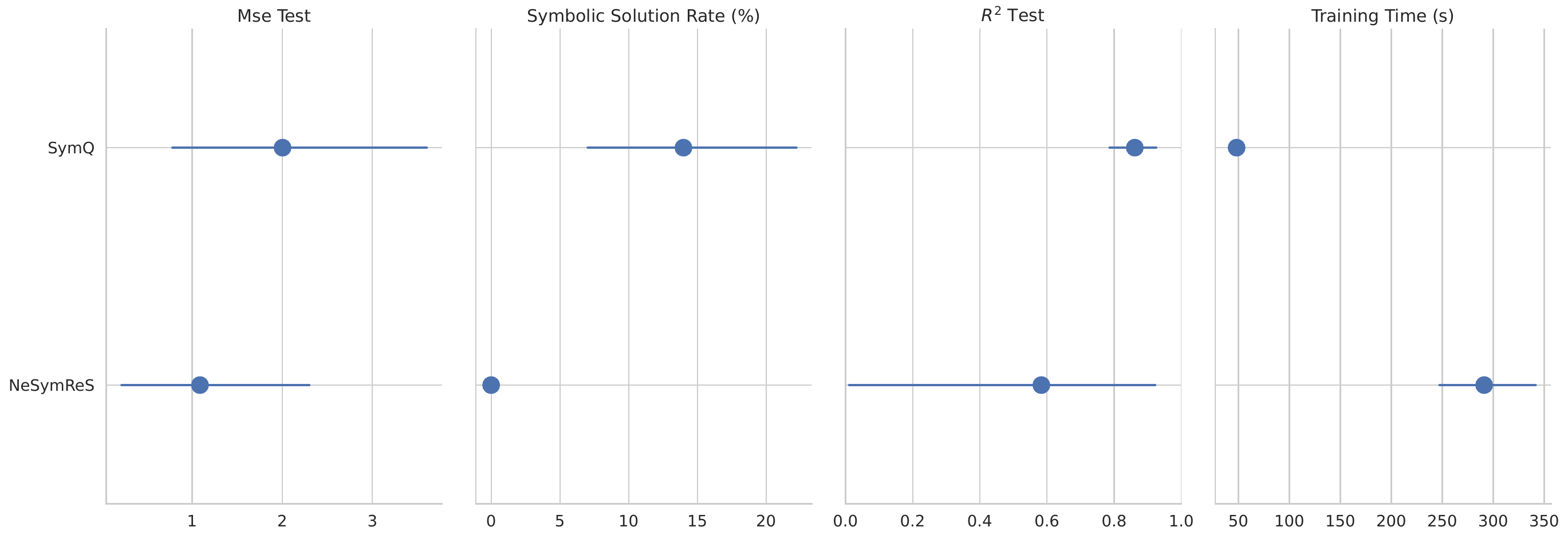}
\caption{Evaluation on SRBench. As a pretrained symbolic regression model, we compared the performance of our model with NeSymReS. To make fair comparison, we have scaled up the model to three independent variables and use the frozen weights of the Set Transformer provided by NeSymReS. The models are evaluated on 50 Feynman equations with up to three independent variables and 14 Strogatz equations. It can been seen that Sym-Q has superior ability in recovering the equation and data fitting. As we do not involve transformer as decoder, the inference time is significantly lower.}
\label{fig: srbench}
\end{figure}



\end{document}